\documentclass[10pt,twocolumn,letterpaper]{article}

\usepackage{tikz}
\usepackage{cvpr}
\usepackage{times}
\usepackage{epsfig}
\usepackage{graphicx}
\usepackage{amsmath}
\usepackage{amssymb}
\usepackage{gensymb}
\usepackage{subcaption}

\usetikzlibrary{shapes.geometric,calc}

\cvprfinalcopy 

\def\cvprPaperID{****} 

\begin{document}

\title{Automatic 3D Reconstruction for Symmetric Shapes}

\author{Atishay Jain\\
Stanford University\\
{\tt\small atishay@stanford.edu}
}

\maketitle

\begin{abstract}
Generic 3D reconstruction from a single image is a difficult problem. A lot of data loss occurs in the projection. A domain based approach to reconstruction where we solve a smaller set of problems for a particular use case lead to
greater returns. The project provides a way to automatically generate full 3-D renditions of actual symmetric images that have some prior information provided in the pipeline by a recognition algorithm. We provide a critical analysis on how this can be enhanced and improved to provide a general reconstruction framework for automatic reconstruction for any symmetric shape.
\end{abstract}

\section{Introduction}

Reconstruction of a 3D object from a single image is a challenging problem. The camera matrix which has 11 degrees of freedom is difficult to obtain with minimal user input. Even after obtaining the camera matrix, the job is half done as the real 3D object is still known to a projective ambiguity. The projective transform itself can have 8 degrees of freedom which we have no way of figuring out from the image. Despite this, most living organisms are perfectly capable of estimating a lot of information about the object with a single eye. A long standing goal of the vision community has been to list down the wide set of rules that the brain uses to identify objects and to apply them to an image reconstruction as constraints that can help get to a reconstruction level closer to the human brain. Despite failing at various optical illusions, a system that matches the skill level of the human eye would be a great leap in technology. \par
\begin{figure}
\begin{center}
   \includegraphics[width=0.8\linewidth]{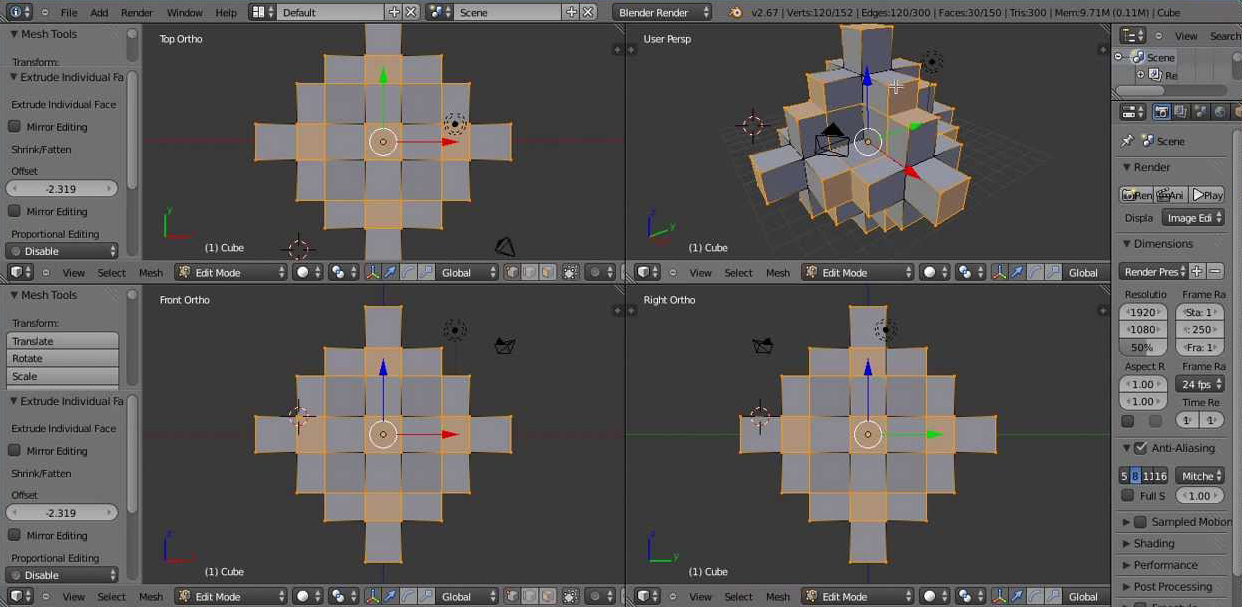}
\end{center}
   \caption{3D modeling in Blender. The representation of a 3D object in 2D is overwhelming.}
\label{fig:blender}
\end{figure}

Considering the difficulty in reconstruction, solutions to a subset of the problem tied to specific domains has been a middle ground. A big use case of reconstruction is modeling. A reconstruction which is close to the real object can be used by artists as props in 3D scenes used in games and movies. Traditionally modeling in 3D is also a very difficult problem. Representation of a full 3D scene in the available memory as well as viewing this as user is challenging. Figure \ref{fig:blender} shows the interface of a 3D modeling tool that the designers use. There is desire to model the objects in 2D that can be converted to 3D without going through the difficult task of trying to make sense of the object across many camera views. \par
\section{Previous Work}
Approaches to this problem have traditionally been in two broad directions. The first direction is starting with the camera parameters and then calibrating the camera such that more data is available for the calculation~\cite{Chellali}. In this case you use a specially crafted calibration rig that is placed in the same image alongside the object to be measured which provides all the information needed to transfer the data like the cross ratios to calculate the actual object. \par
The second approach involves taking an assumption of a model and then treat this as a fitting problem where we try to fit the observed data to our model. The traditional challenge to this approach has been an identification of the model to fit to a given data set. With advances in recognition via deep learning, as the classification and detection problem is getting easier, this approach has been gaining traction in the community and a lot of new approaches have been proposed. These approaches range from trying to fit at abstract renditions of the real object like hand drawn sketches~\cite{sketch} to full 3D models~\cite{model}. \par
Two such fitting approaches that use real images inside the fitting problem are 3-Sweep~\cite{3Sweep} and its extension D-Sweep~\cite{DSweep}. These come into a subset of the above fitting approach called Sweep Based Modeling that has been extensively studied and used in 3D solid object modeling~\cite{SolidModeling}. Both these approaches require manual input from the user, for example clicking an selecting the cross section and depth in 3-Sweep while fitting the cross section and selecting depth in the D-Sweep version. \par
This paper approaches this fitting problem from another direction. Say a recognition system recognizes the bounding box of the object as well as provide some basic information about the object(like it is a extruded circle along y-axis), can we automatically fit a 3D object to the given 2D data using the given information. \par
The following sections describe geometrical information that we can obtain from the image and therefore use for reconstruction as well as the setup that is used to extract this data from the image and the corresponding steps performed for a generic reconstruction with the corresponding results and associated issues.

\section {Problem Statement}
\begin{figure}
\begin{center}
   \includegraphics[width=0.8\linewidth]{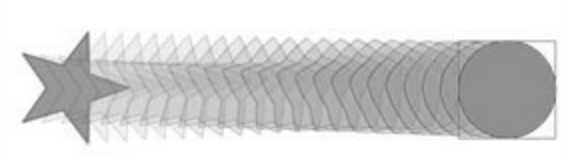}
\end{center}
   \caption{Onion skinning in animation}
\label{fig:onion}
\end{figure}

Given an image we need to identify the surface and the extrusion details from the image so that the object can be reconstructed. The objects targeted for reconstruction are those that can be created via extrusion of a 2D shape along an axis in the third dimension. The extruded object can have a lot of transforms during the process. We will discuss scaling and translation though the concept can be extended to include any transform like rotation and even morphing. The concept of a 2D shape morphing across another dimension preserving the details like texture has been widely studied\cite{morphing} and used in the field on animation where the third dimension is time. Designers are familiar with concepts like onion skinning(Figure \ref{fig:onion}) that show intermediate frames of an animation. In extrusion, the third dimension is the visible 3D space that has been projected onto the image. For the purpose of this project, we assume that there is a recognition piece in the pipeline that identifies the approximate bounding box of the object to be detected, and provides the bounding box of the object corrected by the angle of rotation of the plane of extrusion and the shape of the plane of extrusion (e.g. circle/square).

The solution to the generic extrusion based reconstruction can be broken into the following parts:

\subsection{Plane identification}
The first fitting problem in this form of reconstruction is fitting the cross-section frame of extrusion onto the image. Even though the learning algorithm can give an approximate bounding box of the frame, the actual frame location in 3D space plays a very important role in the determination of extrusion. If this frame has a very good fit, extruding the frame starting with a normal would yield great results. \par
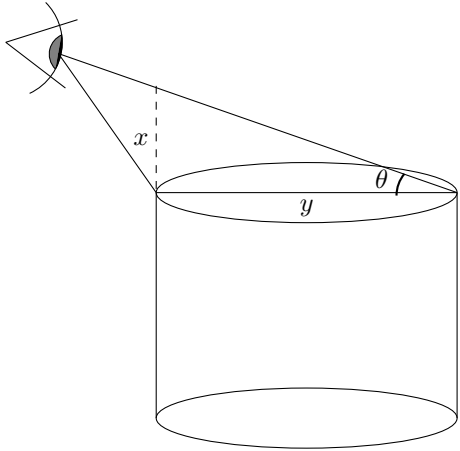
\begin{figure}
\centering
\begin{tikzpicture}
\pgfmathsetmacro{\eyeSize}{1}
\pgfmathsetmacro{\ex}{-4}
\pgfmathsetmacro{\ey}{2}
\pgfmathsetmacro{\eRot}{-10}
\pgfmathsetmacro{\eAp}{-55}
\draw[rotate around={\eRot:(\ex,\ey)}] (\ex,\ey) -- ++(-.5*\eAp:\eyeSize)
     (\ex,\ey) -- ++(.5*\eAp:\eyeSize);
\draw (\ex,\ey) ++(\eRot+\eAp:.75*\eyeSize) arc (\eRot+\eAp:\eRot-\eAp:.75*\eyeSize);

\draw[fill=gray] (\ex,\ey) ++(\eRot+\eAp/3:.75*\eyeSize) 
  arc (\eRot+180-\eAp:\eRot+180+\eAp:.28*\eyeSize);

\draw[fill=black] (\ex,\ey) ++(\eRot+\eAp/3:.75*\eyeSize) 
  arc (\eRot+\eAp/3:\eRot-\eAp/3:.75*\eyeSize);

\draw (-2,0) -- (2,0);
\draw (0,0) ellipse (2 and 0.4);
\draw (0,-3) ellipse (2 and 0.4);
\draw (-2,0) -- (-2,-3);
\draw (2,0) -- (2,-3);

\draw (-3.3,1.85) -- (2,0);
\draw (-3.3,1.85) -- (-2,0);
\draw [dashed] (-2, 1.42) -- (-2,0);

\draw (0, -0.2) node{$y$};
\draw (-2.2, 0.71) node{$x$};
\draw (1, 0.18) node{$\theta$};
\draw [thick] (1.3, 0.25) arc (140:180:0.45);
\end{tikzpicture}
\caption{Camera location assumption}
\label{fig:angle}
\end{figure}

There are three main challenges in fitting this frame:
\begin{figure}
\centering
        \resizebox {0.8\columnwidth} {!} {
\begin{tikzpicture}
\pgfmathsetmacro{\eyeSize}{1}
\pgfmathsetmacro{\ex}{-4}
\pgfmathsetmacro{\ey}{2}
\pgfmathsetmacro{\eRot}{0}
\pgfmathsetmacro{\eAp}{-55}
\draw[rotate around={\eRot:(\ex,\ey)}] (\ex,\ey) -- ++(-.5*\eAp:\eyeSize)
     (\ex,\ey) -- ++(.5*\eAp:\eyeSize);
\draw (\ex,\ey) ++(\eRot+\eAp:.75*\eyeSize) arc (\eRot+\eAp:\eRot-\eAp:.75*\eyeSize);

\draw[fill=gray] (\ex,\ey) ++(\eRot+\eAp/3:.75*\eyeSize) 
  arc (\eRot+180-\eAp:\eRot+180+\eAp:.28*\eyeSize);

\draw[fill=black] (\ex,\ey) ++(\eRot+\eAp/3:.75*\eyeSize) 
  arc (\eRot+\eAp/3:\eRot-\eAp/3:.75*\eyeSize);

  \node[cylinder,draw=black,aspect=0.7,minimum height=1.7cm,minimum width=1.5cm,shape border rotate=90] at (0, 0) (A) {Top};
  \node[cylinder,draw=black,aspect=0.7,minimum height=1.7cm,minimum width=1.5cm,shape border rotate=270] at (0, 4) (B) {Bottom};
  \node[cylinder,draw=black,aspect=0.7,minimum height=1.7cm,minimum width=1.5cm,shape border rotate=270] at (3, 2) (C) {};
  \node[fill,text width=1.2cm, minimum height=1.6cm, fill=white] at (3.0,1.94) {None};
\draw [dashed] (-3.3,2) -- (2.3,2);
\draw [dashed] (-3.3,2) -- (0,3.2);
\draw [dashed] (-3.3,2) -- (0,0.8);
\end{tikzpicture}
}
\caption{Visibility of the plane based on the location of the view}
\label{fig:visibility}
\end{figure}
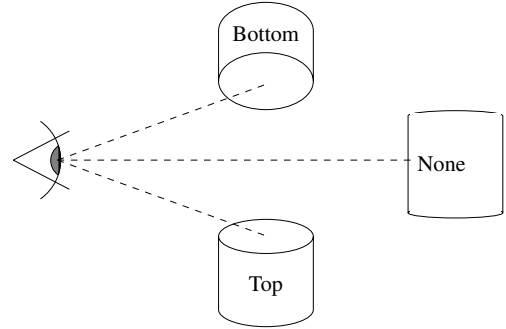
\begin{itemize}
   \item \textbf{Occlusion:} Based on the angle of view, parts of the frame may be occluded from view(Figure \ref{fig:visibility}). It may happen that the entire frame is occluded and only a curve representing the intersection of this frame with the rest of the object is visible. Additionally, there can be external occlusion where outside objects obstruct the view.
   \item \textbf{Projection:} The projective transform leaves an ambiguity in the actual structure of the plane. The shape of the actual plane can be determined by prior knowledge which the detection algorithm earlier in the pipeline should provide. D-Sweep relies on the user to provide the shape while 3-Sweep uses two sweeps to estimate this shape with the simple assumption of it being a circle or a rectangle. In most practical cases the projection starts perpendicular to this view. This assumption can help in further simplification of the problem. The estimation of this projection can give a very good idea about the field of view(See Figure \ref{fig:angle}). $$ \tan \theta = \frac{x}{y} $$.
   Note that, this data only provides information about the plane or the cross-section that was attempted to be fit and not the extrusion.
   \item \textbf{Bevel:} In many cases where the bevel of the object is smooth, this frame may not even have a visible edge. Alternatively, the bevel could be stepped that has too much noise for fitting to work properly. There could be multiple types of noise in the measurement, including uneven lighting, shadows texture, background etc, than can make fitting even tougher. \par
\end{itemize}

   An approach that uses both the color/spatial information as well as a heuristic based on the geometry can yield good fitting to the plane. \par
   For the special case where $\theta = 0$, this plane reduces to a straight line. Apart from external occlusion, the challenges above cease to exist. The detection also becomes one dimensional and is a simpler problem. The downside of having $\theta = 0$, is that we do not get to fit the cross section passed by the previous case and all cross sections (see Figure \ref{fig:rectangle}) can be fit into the object. The feedback loop into the previous stages is lost.
\begin{quote}
\begin{center}
For the rest of the sections we assume that $\theta = 0$.
\end{center}
\end{quote}
\subsection{Edge matching and extrusion}
After plane identification fitting the edges of the object to the boundaries of extrusion would complete the 3-D object. The face detected above could be extruded with any sort of transformation across the Z-dimension. There is also a projective ambiguity in play. Therefore we can never fully estimate this angle. This projective ambiguity can only be rectified based on some assumptions. An approach is to start with some hypothesis and try fitting the model based on those sets of parameters. Pick the best fit amongst the various hypothesis. \par
The basic assumptions that can lead to good results include assuming that the object has minimal scaling. Tapering objects are rarer than straight ones. Therefore more weight-age should be given to a model which is rotated and has projective transforms over scaling models across the image. Also weighting information from multiple objects in the same image as well as prior knowledge can provide good results. We can employ weighting to measure the ground plane's angle to the camera from multiple objects. In most of the cases the plane of the object is almost perpendicular to the plane of extrusion. Therefore additional weight can be given to the assumption. The possibility of a sudden change in angle is rare. In 3-Sweep Chen et al. look at only $\pi/3$ radians from the previous location in search for the next extrusion point. \par
Another part of this process is matching rotation and morphing. For objects that have sharp edges (anything with solid bevels), the edges are also available as a change in gradient in the image. The default behavior of lighting is such that the edge is visible. These edges in the middle of the object can help decipher the rotation of the frame across the length of the object as well as shape morphing. We can use consistency across frames as the outliers removal criteria. \par
A lot of use cases do not require that complexity. Again, if the user focus is capturing only the desired object, the default behavior is to capture the image with the object at the center of the screen and parallel to the frame of the image. In this case, the tapering does not exist and mapping becomes simpler. That approach is the best to get the desired texture as the object is able to take the maximum amount of space on the image and therefore most of texture that can be viewed is available if this assumption is made. \par
\begin{quote}
\begin{center}
For the rest of the sections we assume that the image is captured from the front of the object and the tapering can be neglected.
\end{center}
\end{quote}
\subsection{Texture Mapping}
The problem with texture mapping in single image reconstruction is that we don't have major parts of the texture to map to the actual object. This loss of data is irrecoverable. The only assumptions we can make to save grace is that of symmetry. Will the symmetrical assumptions, we can take the the texture from the detection, unscaled this based on the extrusion and camera parameters calculated, repeat to fit the entire object and apply. \par
Cleaning up the texture is a hard problem. There are a lot of issues with the approach that require further work: \par
\begin{itemize}
\item \textbf{Lighting} The lighting can have a significant impact on the textures, especially on objects whose surface is specular. Cleaning spots of light or gradients generated by uneven lighting distribution across the object is a significant task. Add to that the noise and the potential of having multiple lights makes this a hard problem. There is also reflection which needs to be cleaned up for the texture to be smoothly represented everywhere.
\item \textbf{Extremities} Even though we miss data in the real world, we need to make the reconstruction such that the user fails to notice that. This is again a difficult problem. The edges of the parts of the texture extracted may not match end to end and may produce a seam that may not be desired. A workaround that we employ to make this problem less perceptible is to use a mirror of the object when we connect it. This way, the edge matching is performed automatically and apart the surface gradient direction change, there is no perceptible change. A lot of textures are symmetric in nature and do not have perceptible spatial differences. Apart from lighting, there is no difference in the texture if we take this from different parts of the object. The patterns may not be computationally repeatable but visually a lot of patterns are indistinguishable. In these special cases, the texture doubling works smoothly.
\item \textbf{Transparency} Transparency in objects as well the concept of the background color bleeding(an artifact of diffusion/reflection on the image), can make recovery of the texture very difficult.
\item \textbf{Noise/Occlusion} There are multiple sources of noise. Camera inaccuracies like pixelation and graining, atmospheric inaccuracies like mist and fog, surface properties like roughness, dust, uneven reflection, imaging inaccuracies like blur and occlusion of the object plays major roles in making this problem more difficult.
\end{itemize}
Auto-generating a clean texture that can be used at multiple locations would require a lot of prior knowledge about the materials such that the basic properties like specularity of the surface can be determined. A learning system that has this information still needs to figure out solutions to the problems discussed above before a generic solution can be made. We instead focus on a specialized solution. The reconstruction is done in the context of using the model in some setup. If the setup for capturing the image is similar to the scene where the object is going to be used, a lot of these inaccuracies can be neglected. A lot of props that are used in real world are not a focus of the consumer of the 3D scene and automating these inaccuracies in those objects are inconsequential.
\subsection{Pose Fitting}
After the 3D object is generated based on the pose, the task of applying the pose is comparatively easy. The biggest challenge in pose reconstruction is to work around the inaccuracies with the fitting in the earlier steps. Smoothening at some stage would change the object slightly and the effects of the change will be clearly visible. The task of fitting the object into the image is only required if the image has usability outside the generation task. This goal, though is counter intuitive. If the object has usability outside of generation then there would be multiple objects that  present in the scene that may be occluding one another. This makes segmentation and therefore generation difficult. We do not perform the task of pose fitting in this project.
\section{Experimental Setup}
The experimental setup performs the steps as discussed above.
\subsection{Segmentation and Class Identification}
\begin{figure}
\centering
\includegraphics[width=0.4\linewidth]{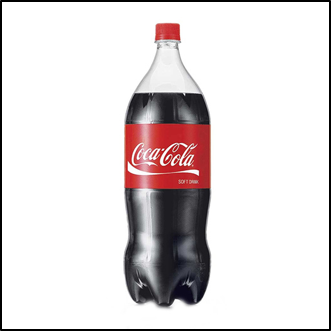}
\includegraphics[width=0.4\linewidth]{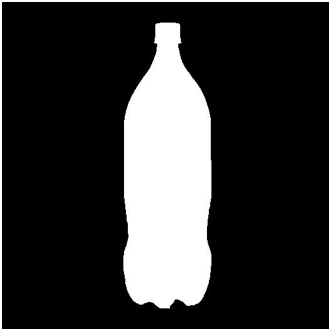}
\includegraphics[width=0.8\linewidth]{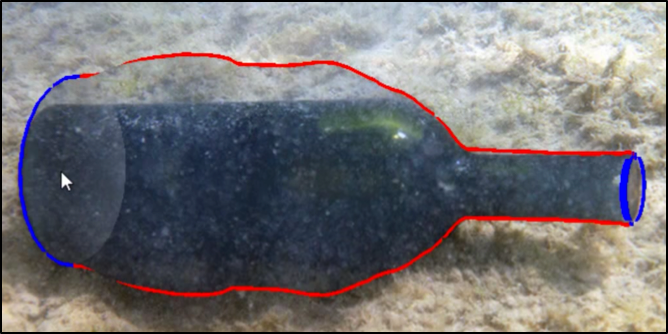}
\caption{From top-left, the input to the algorithm, the output of the segmentation phase, the limitation of 3-Sweep where the object segmentation is difficult.}
\label{fig:segmentation}
\end{figure}

For the experiment, we did not use a learning system that provided this data. We assume that the image size if $512\times 512$ and the object is placed within $(10, 10)$ and $(512, 512)$. We use grabcut\cite{grabcut} to select the object from the image. We take the class parameter as a user input. By default, we assume that the object is cylindrical. Note that the approach is not worse than the manual approaches like 3-Sweep, as edge fitting is tedious and inaccurate. The parameters of the grabcut algorithm are left to default. For better results, segmentation could be applied via a family of parameters and methods and the best results based on the weights obtained by the final pose based on the parameters can be chosen. The results of this phase are shown in Figure \ref{fig:segmentation}.
\subsection{Plane Identification/Profiling} 
\begin{figure}
\centering
\includegraphics[width=0.4\linewidth]{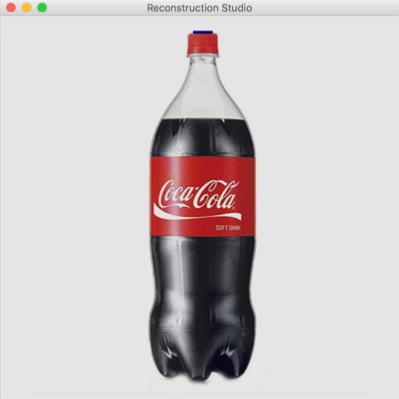}
\includegraphics[width=0.4\linewidth]{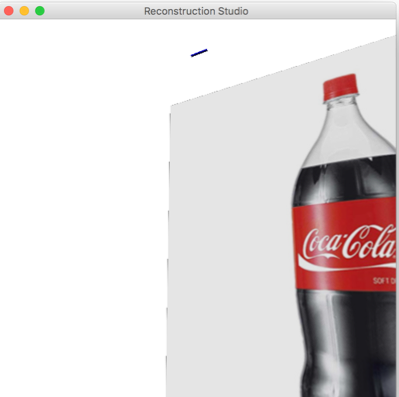}
\caption{The output for the profiling phase of the fitting. We are showing a line that represent a plane to the end user.}
\label{fig:profiling}
\end{figure}

The assumption of the view being the front view with the plane of extrusion being perpendicular to the image frame turns this problem into edge detection. We implement detect the top left corner of the image and move towards right to the corresponding edge on the right to find the other end of this plane. The results of basic profiling degrade as the angle $\theta$ in Figure \ref{fig:angle} gets significantly greater than zero. There are techniques that can be employed to mitigate this problem like looking at a set of points rather than a single one and finding the local maxima for the distance or trying to fit an ellipse to the top of the segmentation result. the implementation is similar to the steps performed by an artist manually tracing the extrusion where the actual image is placed at distance and the surface of extrusion is used drawn by hand and on a 2D frame and fitted to the section on the image. Figure \ref{fig:profiling} shows the output of this phase.

\subsection{Extrusion/Sweeping}
In his phase we we take the two end points of the plane identified and start fitting the edges as we move vertically down. Currently we look for one pixel per line sweep but this method can be improved by using an edge walking algorithm such as marching squares, the two dimensional version of marching cubes\cite{marchingsquares}. This way we can easily handle scale and translation of the plane across the third dimension. For handling rotation and shape morphing, salient points need to be identified in the image and their position needs to be matched during the sweep. Using that approach, we can support almost all types of extrusion. Currently we assume that the image contains a single smooth extrusion. But this extrusion can be removed by corner detection and restarting the fitting and extrusion operation from a detected corner. \par
\begin{figure}
\centering
\includegraphics[width=0.4\linewidth]{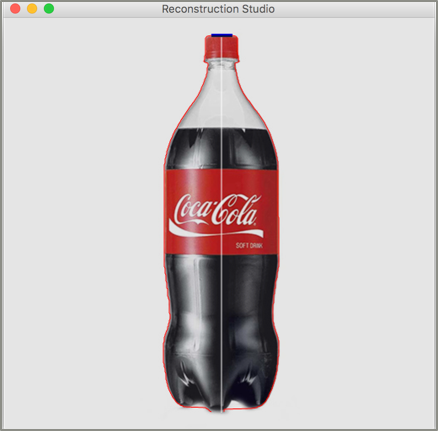}
\includegraphics[width=0.4\linewidth]{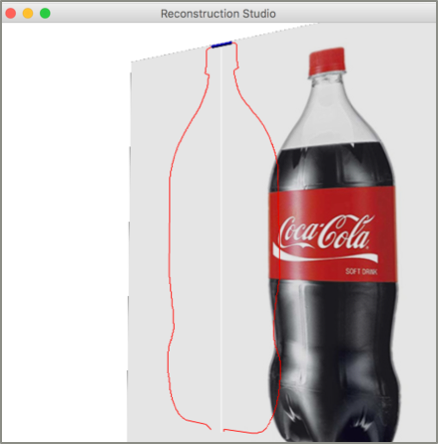}
\includegraphics[width=0.4\linewidth]{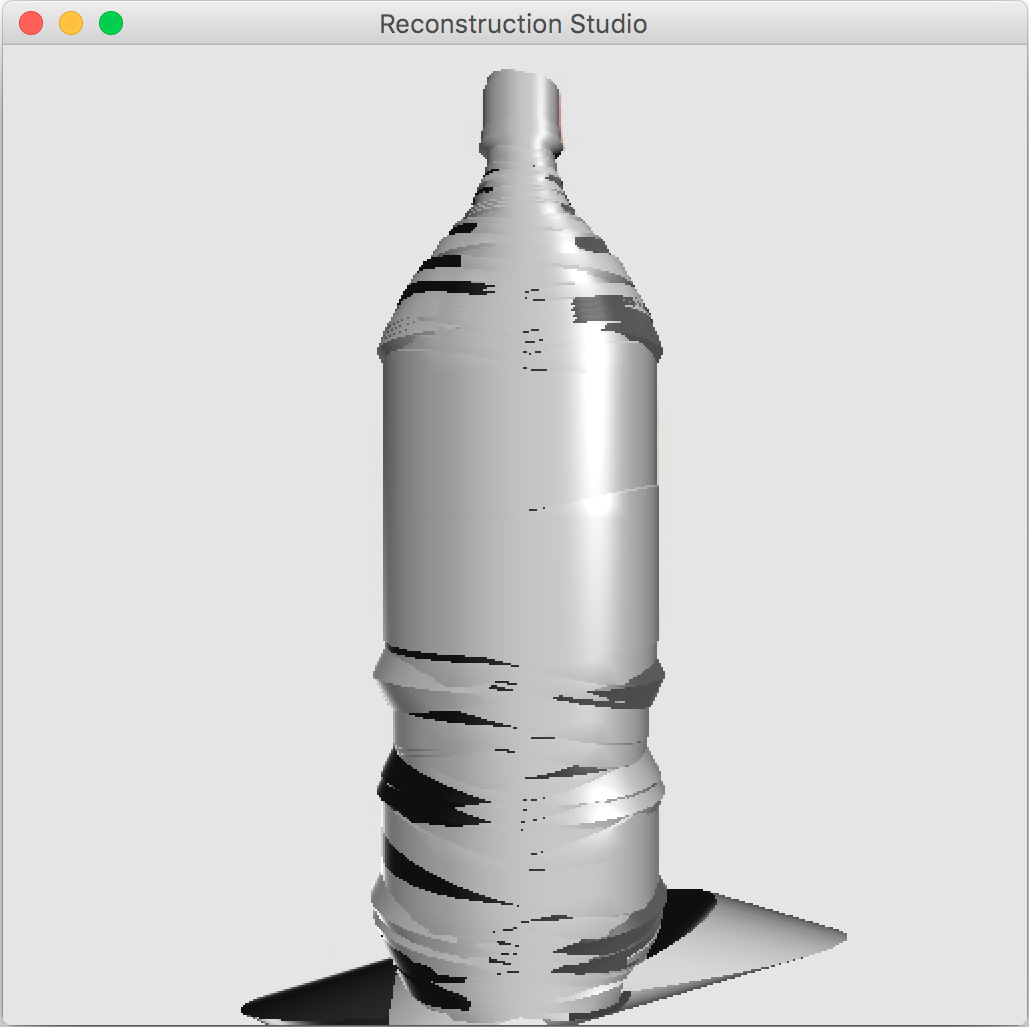}
\includegraphics[width=0.4\linewidth]{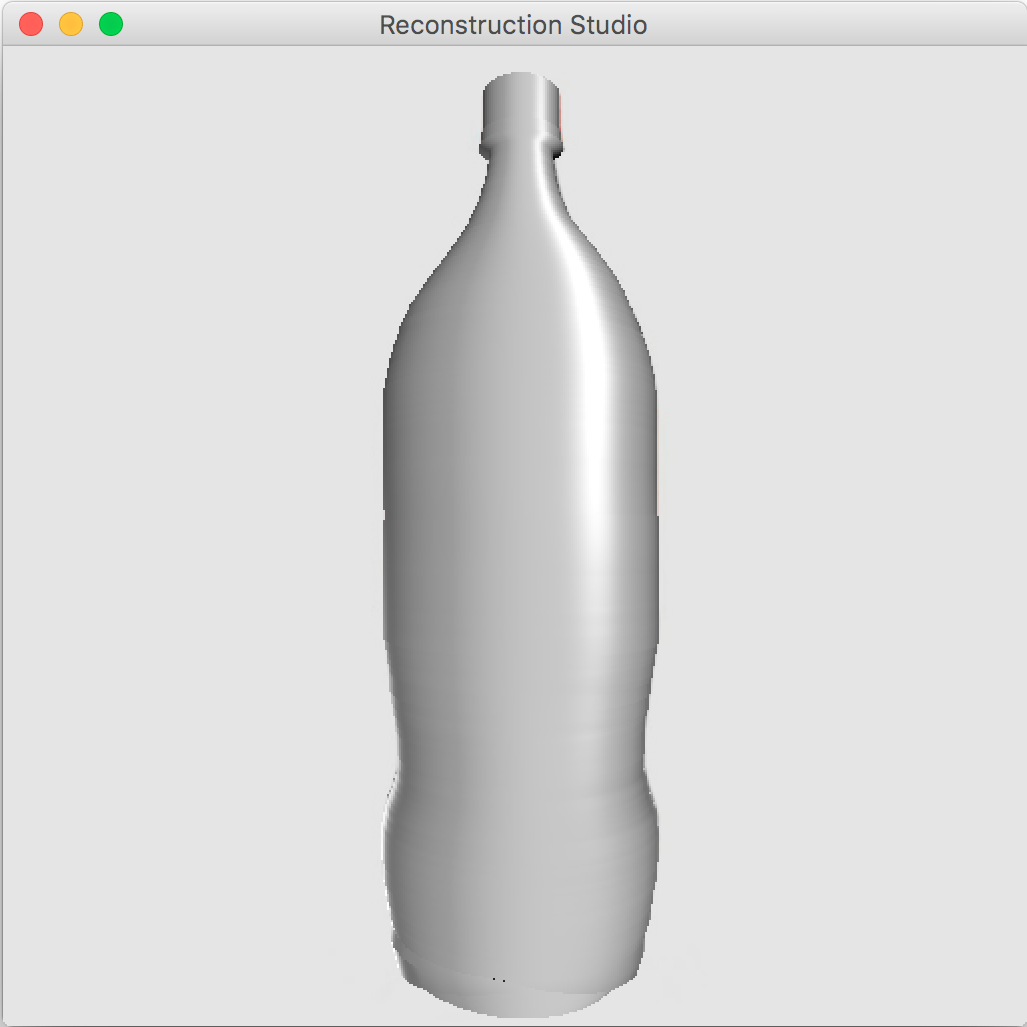}
\caption{From top left, The fitting done in the extrusion phase, the fitting shown from another angle, the output without the smoothening filter, the output with the Savitzky-Golay filter}
\label{fig:sweep}
\end{figure}

There can be pixelation and other issues that can cause significant noise. This noise causes the object to be deformed and not visually appealing. We use the Savitzky-Golay filter\cite{savitzky64} on the center on extrusion to remove the additional noise. The effect of the filter is shown in Figure \ref{fig:sweep}. \par
\begin{figure}
\centering
\includegraphics[width=0.4\linewidth]{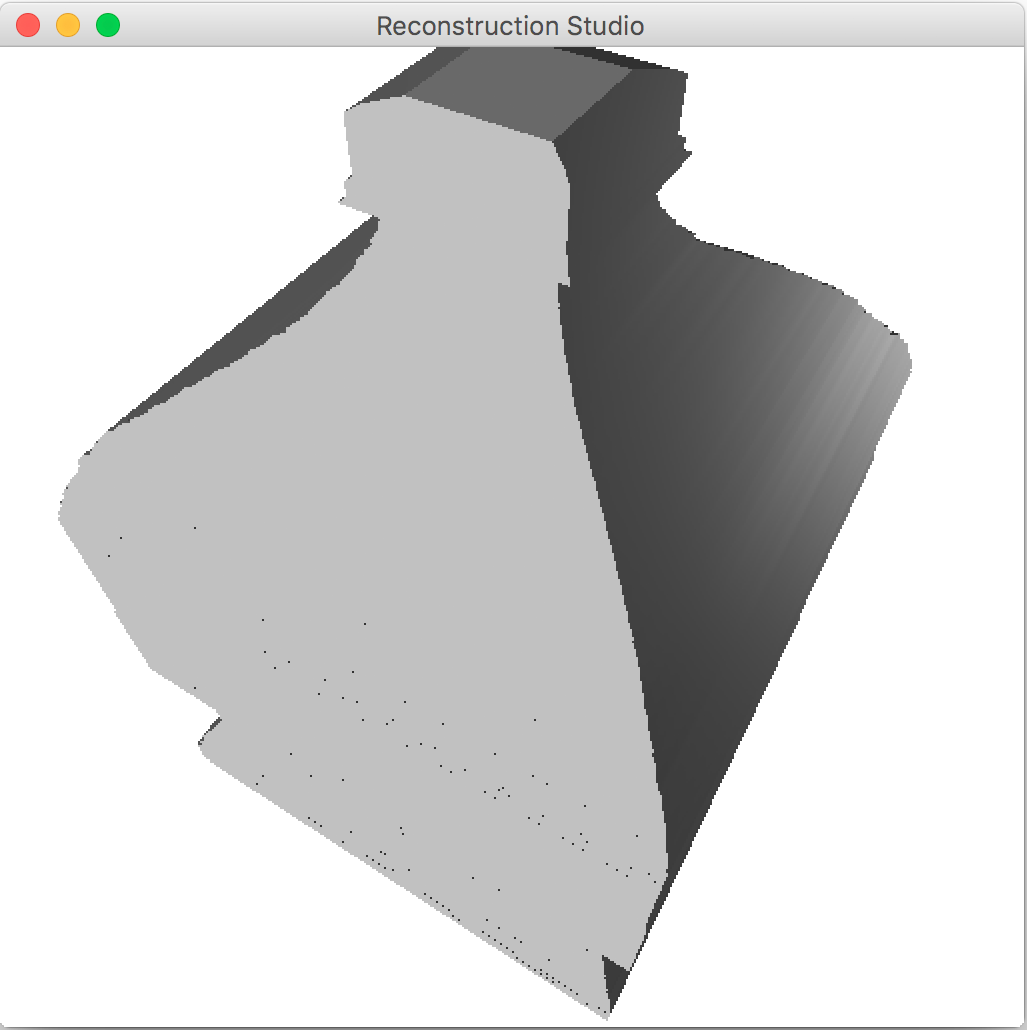}
\includegraphics[width=0.4\linewidth]{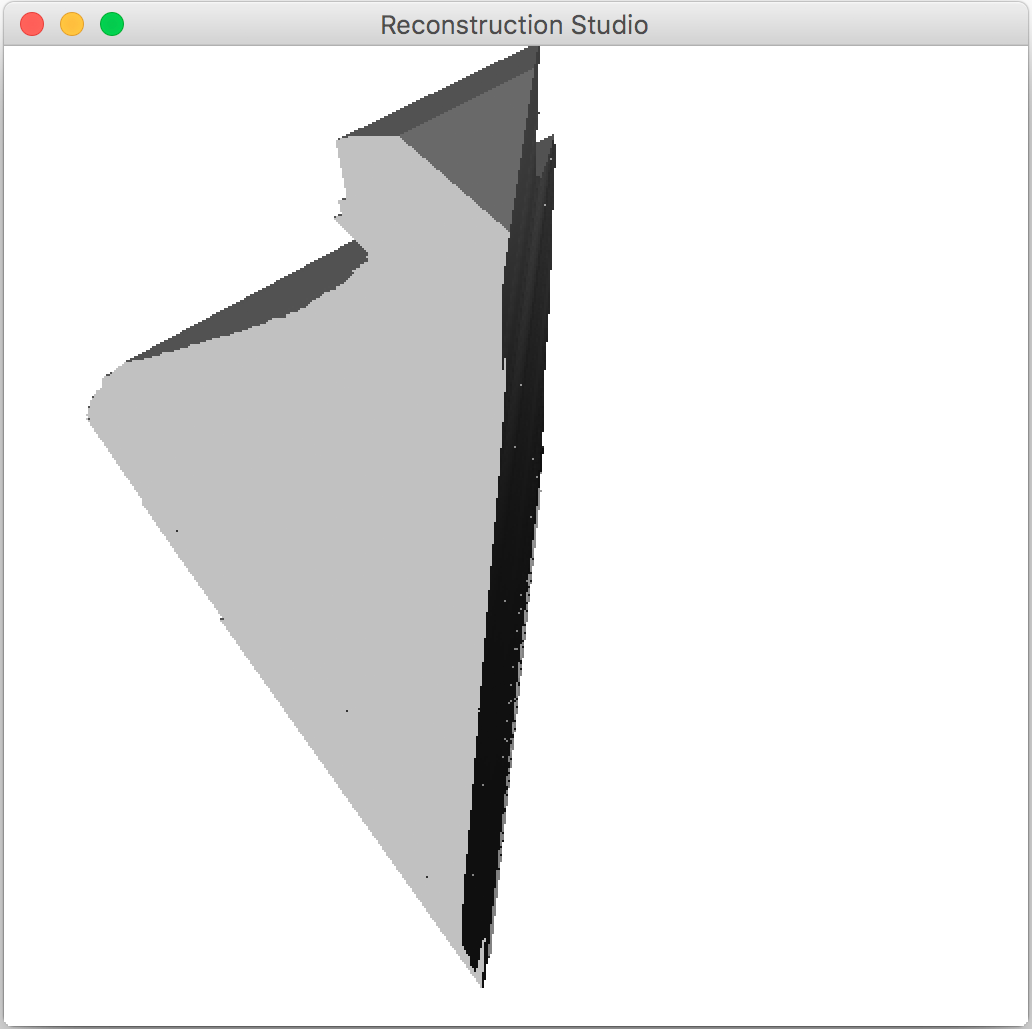}
\includegraphics[width=0.4\linewidth]{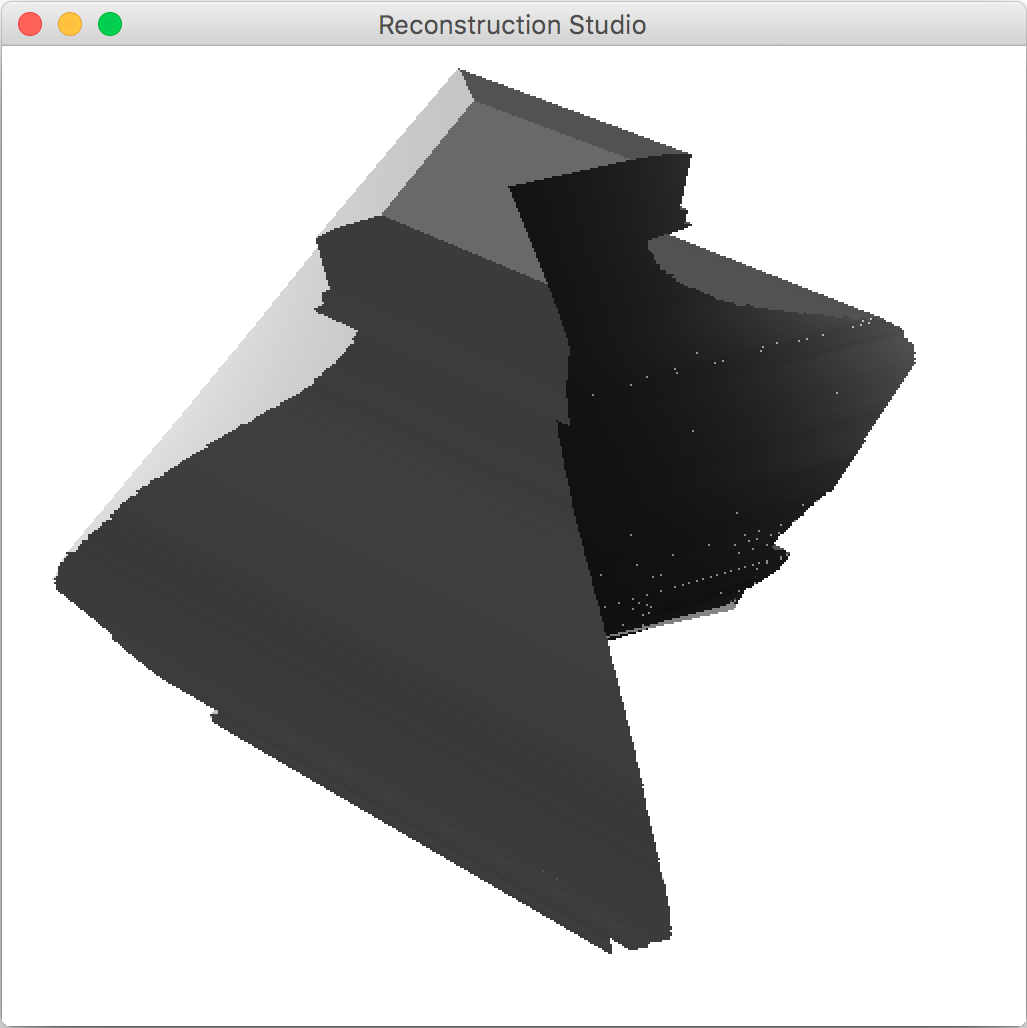}
\includegraphics[width=0.4\linewidth]{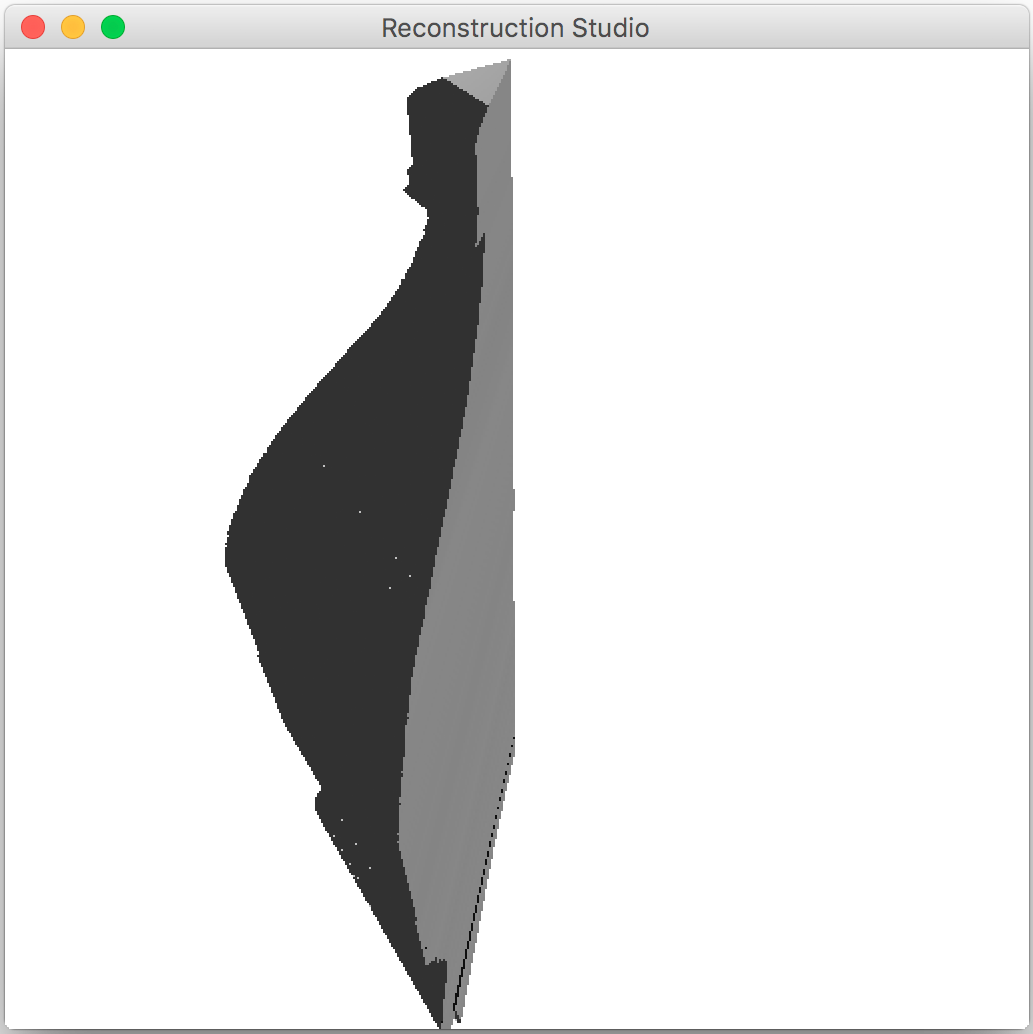}
\caption{Using custom cross sections}
\label{fig:rectangle}
\end{figure}

Note that we do not necessarily need a circular cross section for fitting. We can use any arbitrary shape provided as we assume that the shape is provided to use. We can easily match a square or a triangle cross section, or even custom ones.(See Figure \ref{fig:rectangle}).

\subsection{Texture Mapping} 
\begin{figure}
\centering
\includegraphics[width=0.6\linewidth]{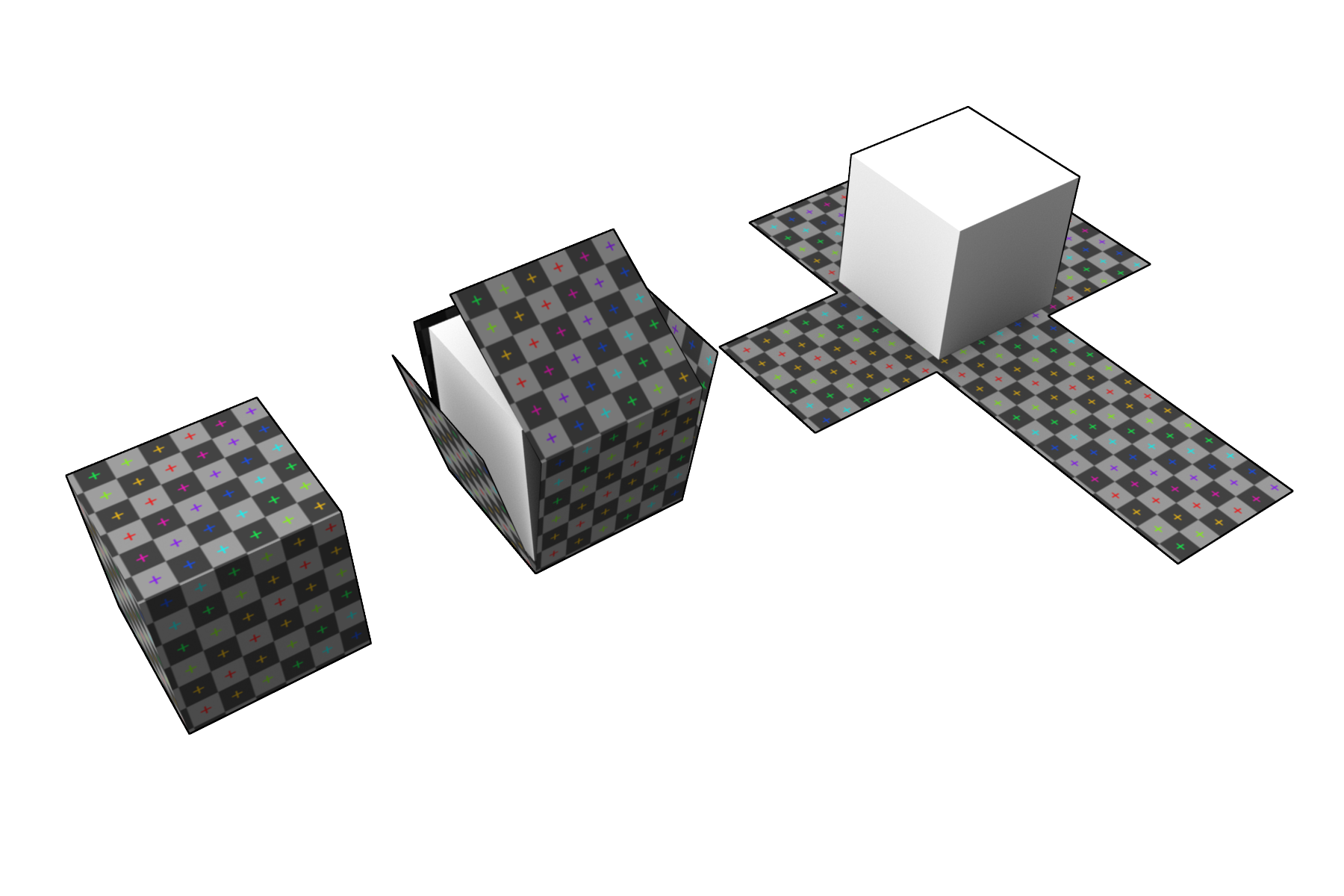}
\includegraphics[width=0.2\linewidth]{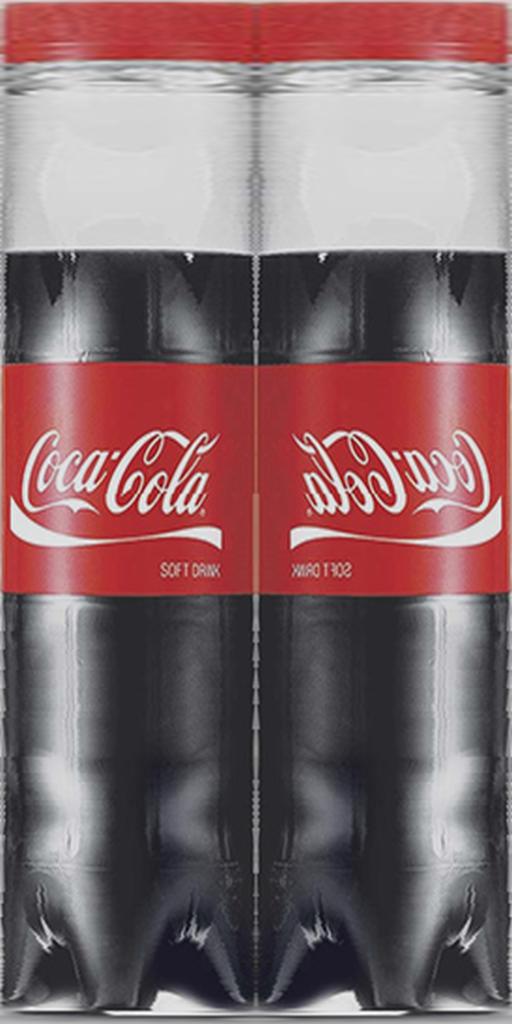}
\caption{Left: Ideal UV unwrapping concept, Right: Cylindrical texture in our experiment}
\label{fig:uv}
\end{figure}

\begin{figure}
\centering
\includegraphics[width=0.4\linewidth]{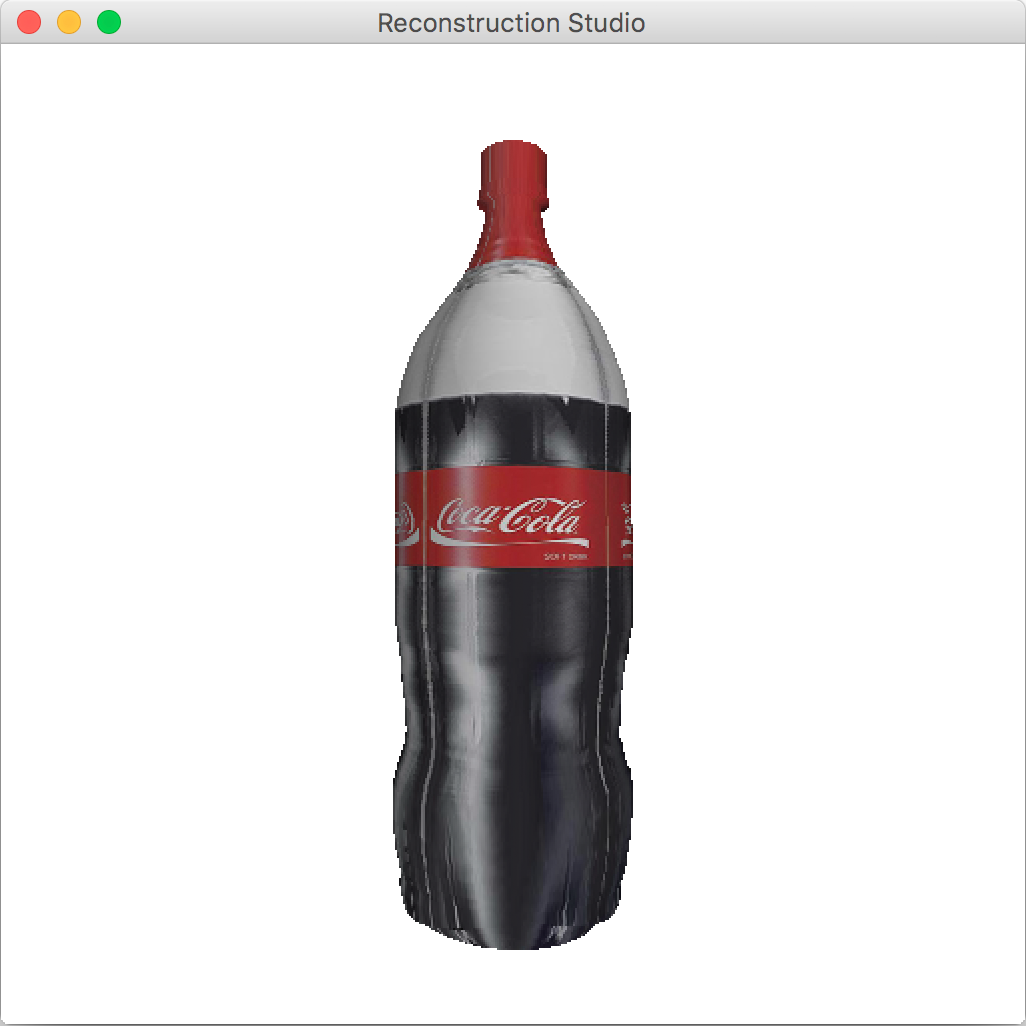}
\includegraphics[width=0.4\linewidth]{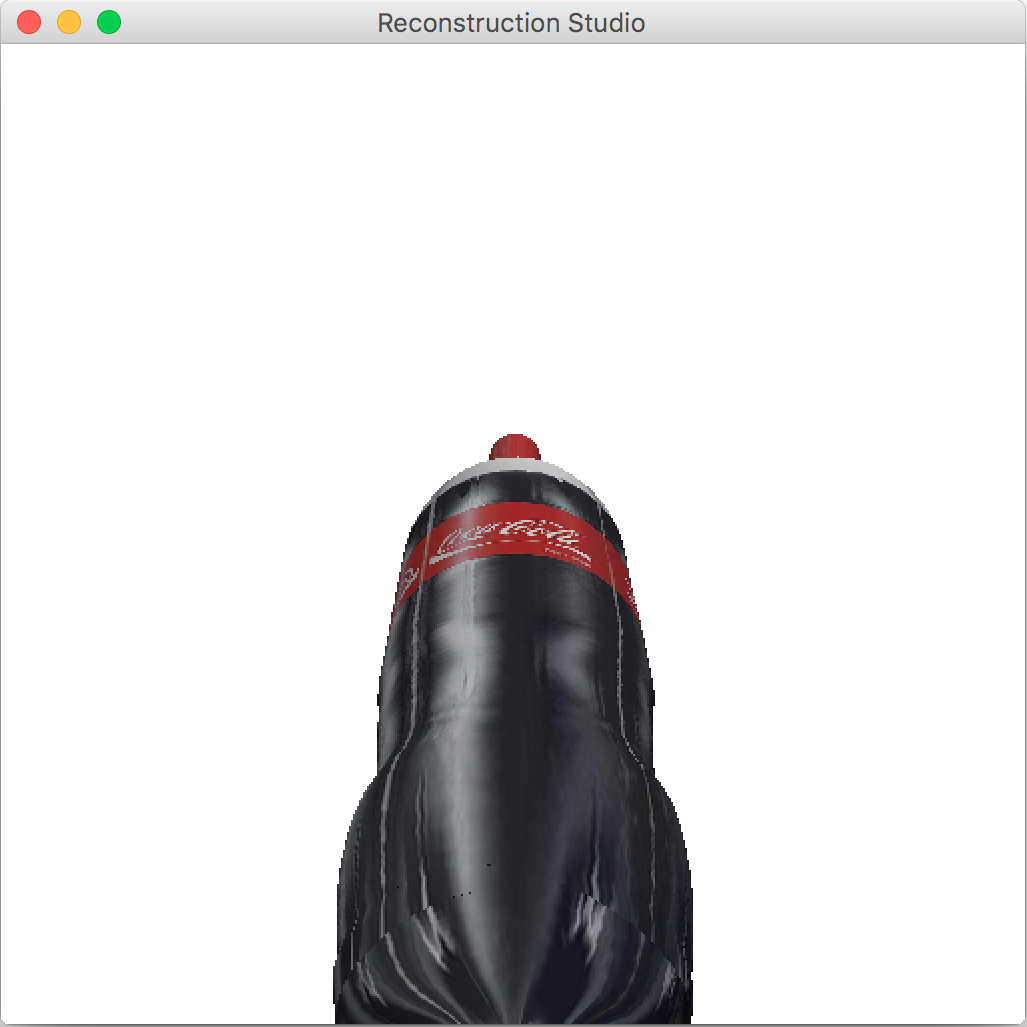}
\caption{Output from two different angles}
\label{fig:final}
\end{figure}

We already have the texture as a part of the image and segmentation mask. The ideal texture mapping in the 3D world requires UV-unwrapping(See Figure \ref{fig:uv}) but we use a simplified cylindrical texture mapping. The cylindrical mapping gives good results if the texture is a repeatable pattern and does not perform as well in other cases. The final output after texture mapping is shown in Figure \ref{fig:final}

\begin{figure}
\centering
\includegraphics[width=0.40\linewidth]{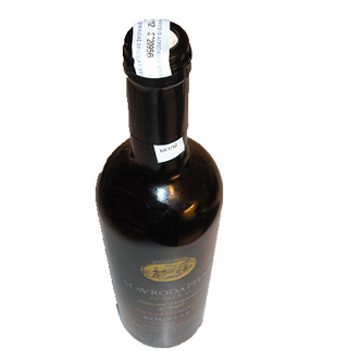}
\includegraphics[width=0.40\linewidth]{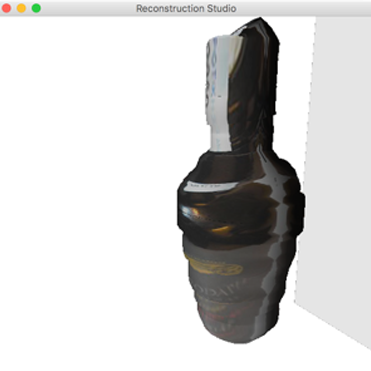}
\caption{Output with tilted image input}
\label{fig:tilt}
\end{figure}
\begin{figure*}
\begin{center}
\setlength{\fboxsep}{0pt}%
   \fbox{\includegraphics[width=0.19\linewidth]{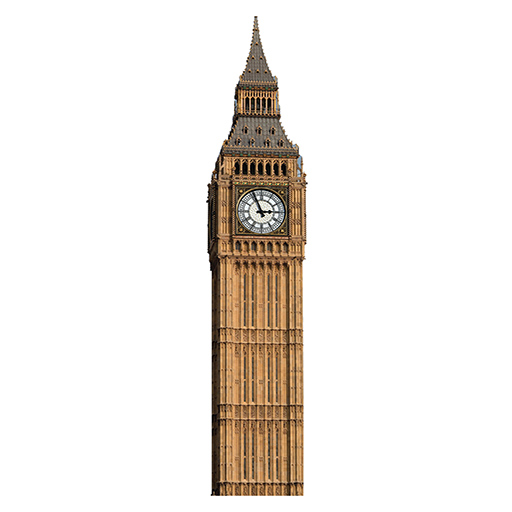}}
   \fbox{\includegraphics[width=0.19\linewidth]{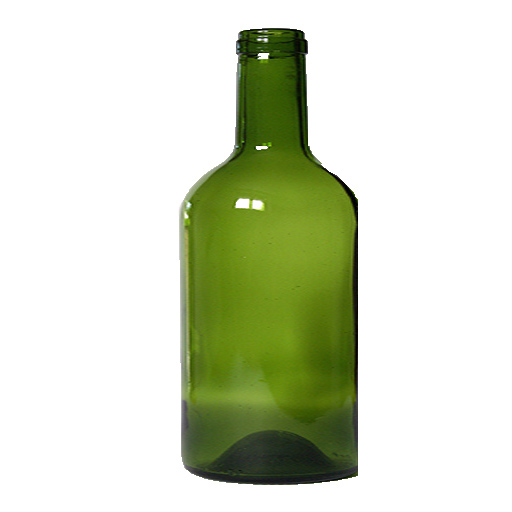}}
   \fbox{\includegraphics[width=0.19\linewidth]{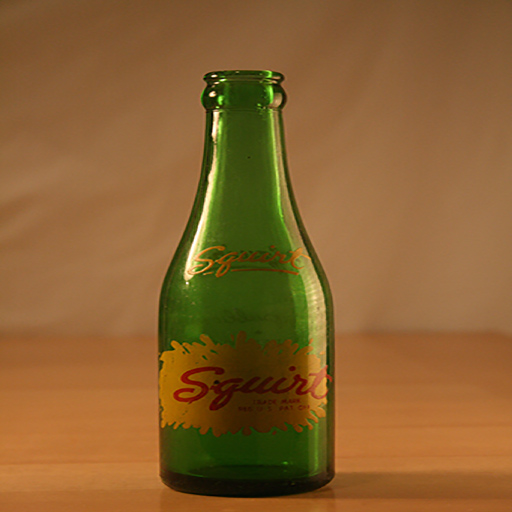}}
   \fbox{\includegraphics[width=0.19\linewidth]{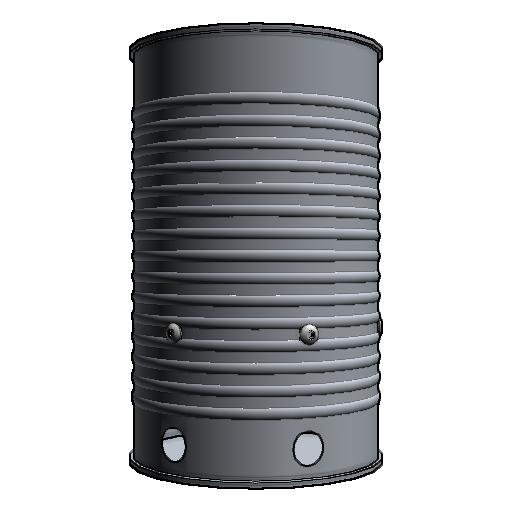}} \\
   \fbox{\includegraphics[trim={3cm 2.9cm 2.2cm 2cm},clip, width=0.19\linewidth]{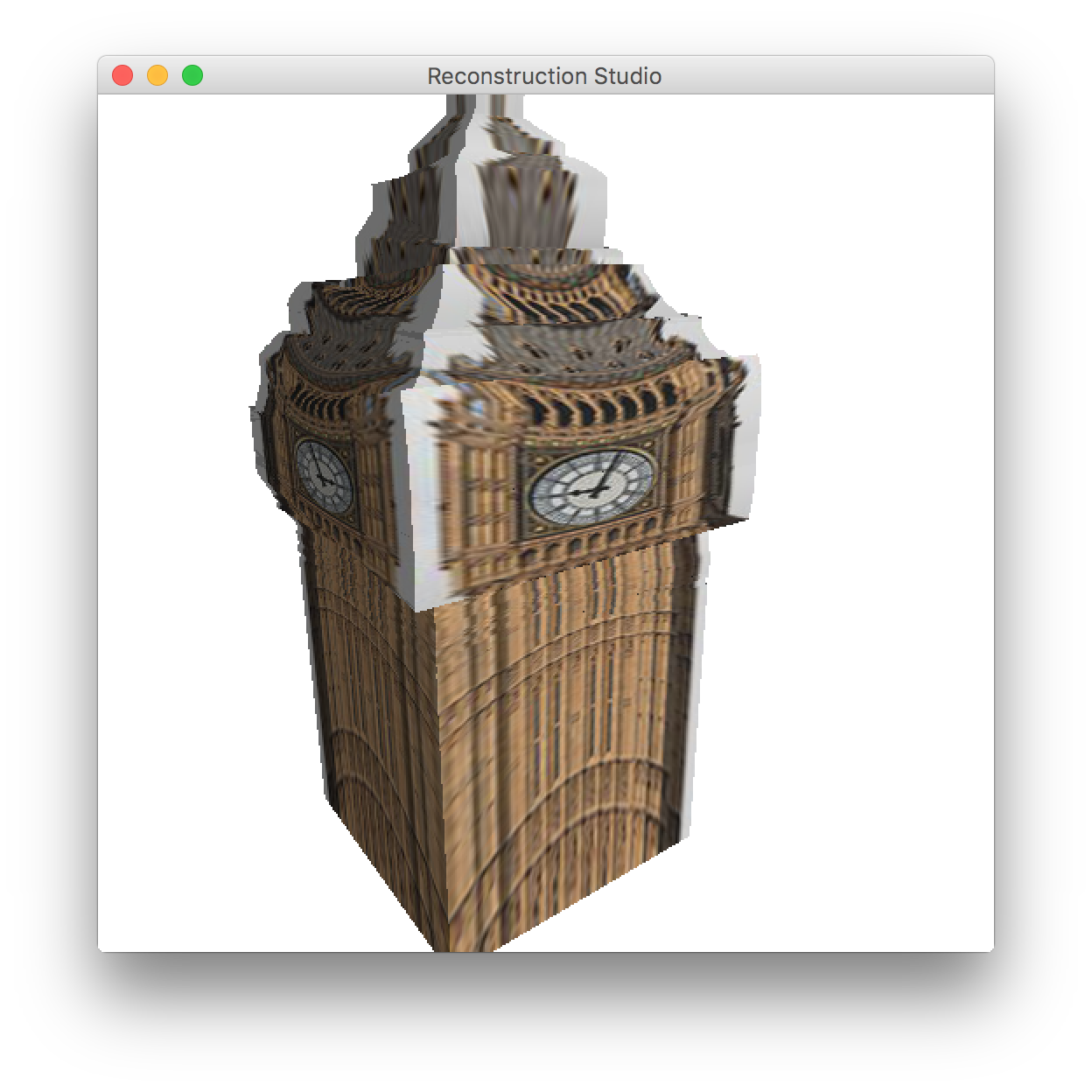}}
   \fbox{\includegraphics[trim={3cm 2.9cm 2.2cm 2cm},clip, width=0.19\linewidth]{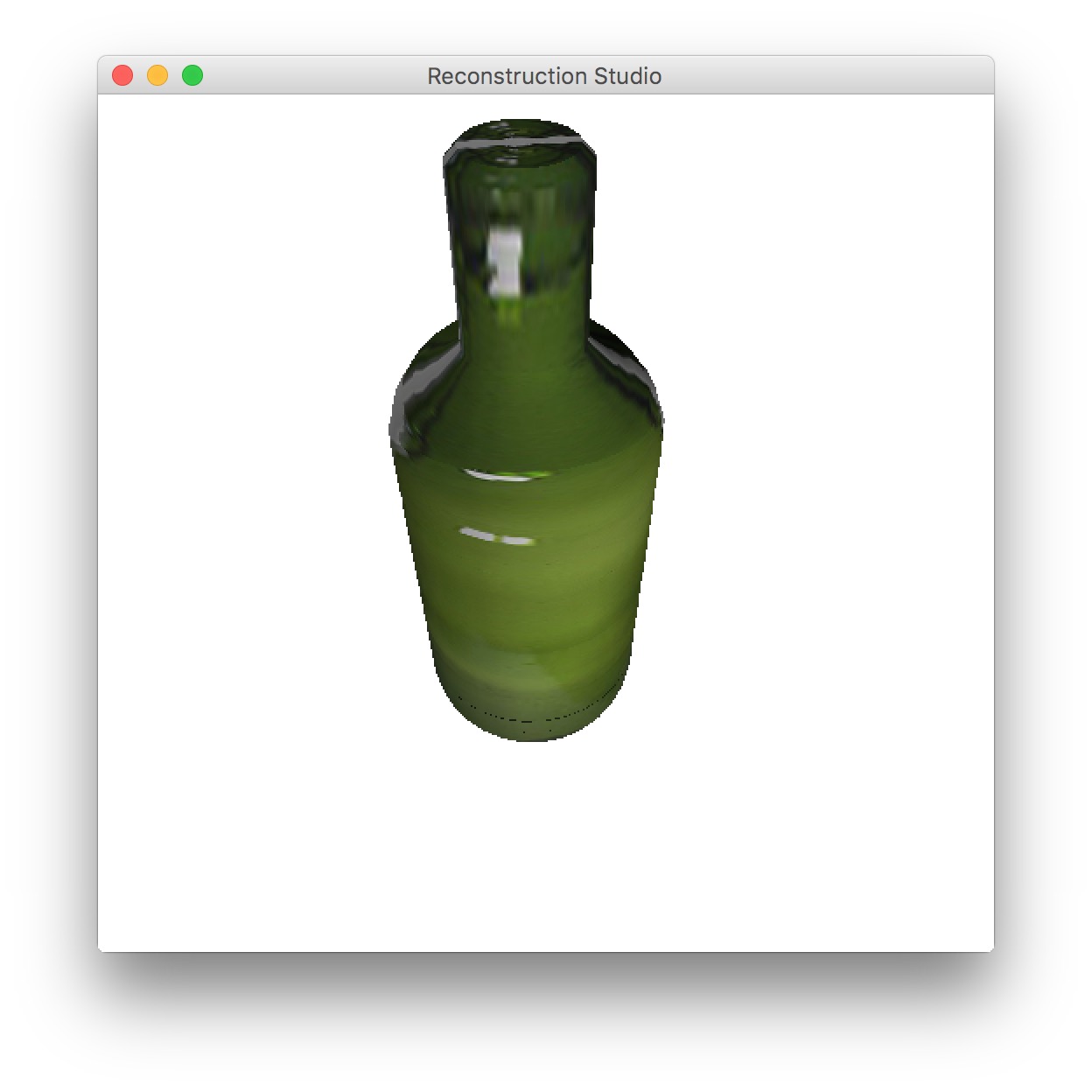}}
   \fbox{\includegraphics[trim={3cm 2.9cm 2.2cm 2cm},clip, width=0.19\linewidth]{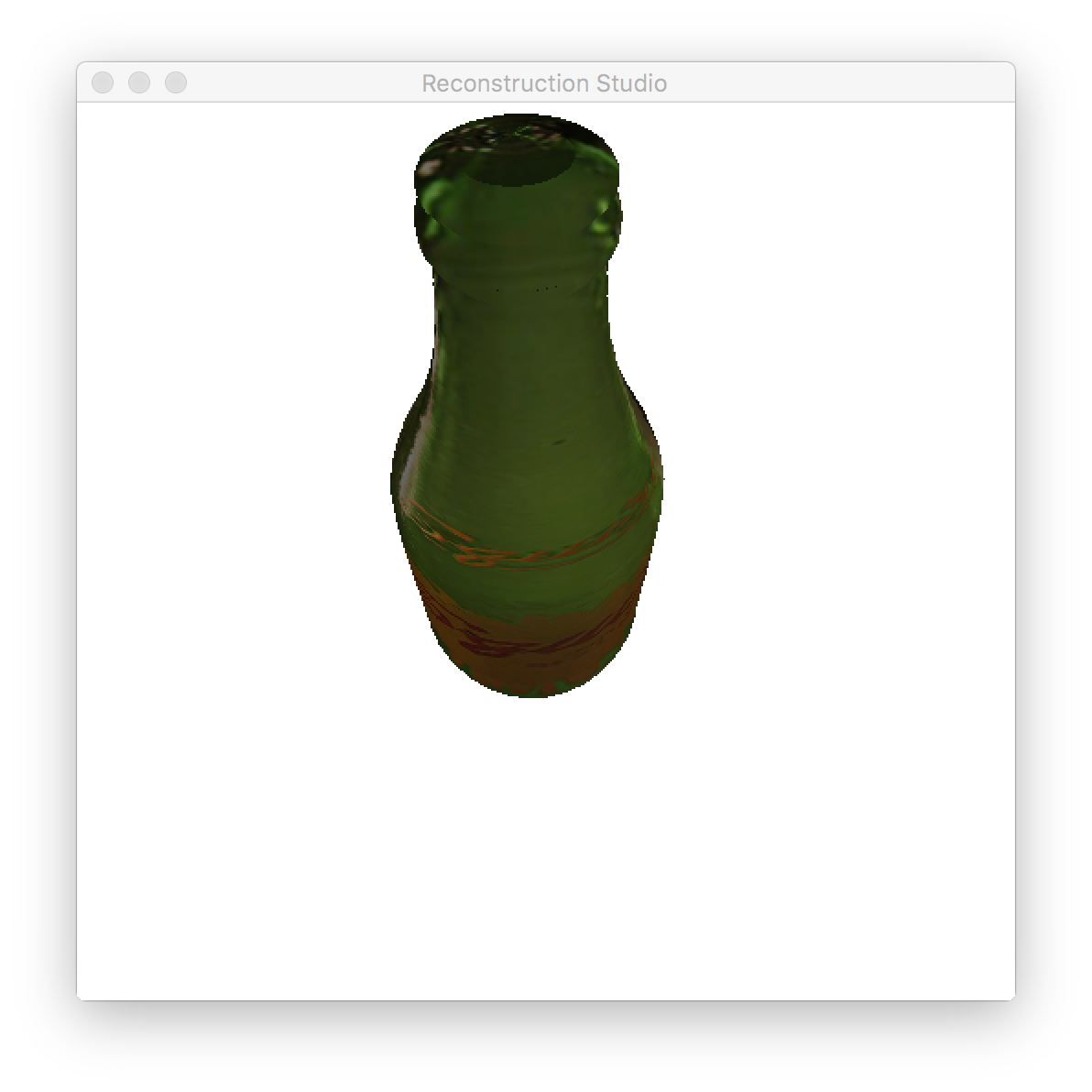}}
   \fbox{\includegraphics[trim={3cm 2.9cm 2.2cm 2cm},clip, width=0.19\linewidth]{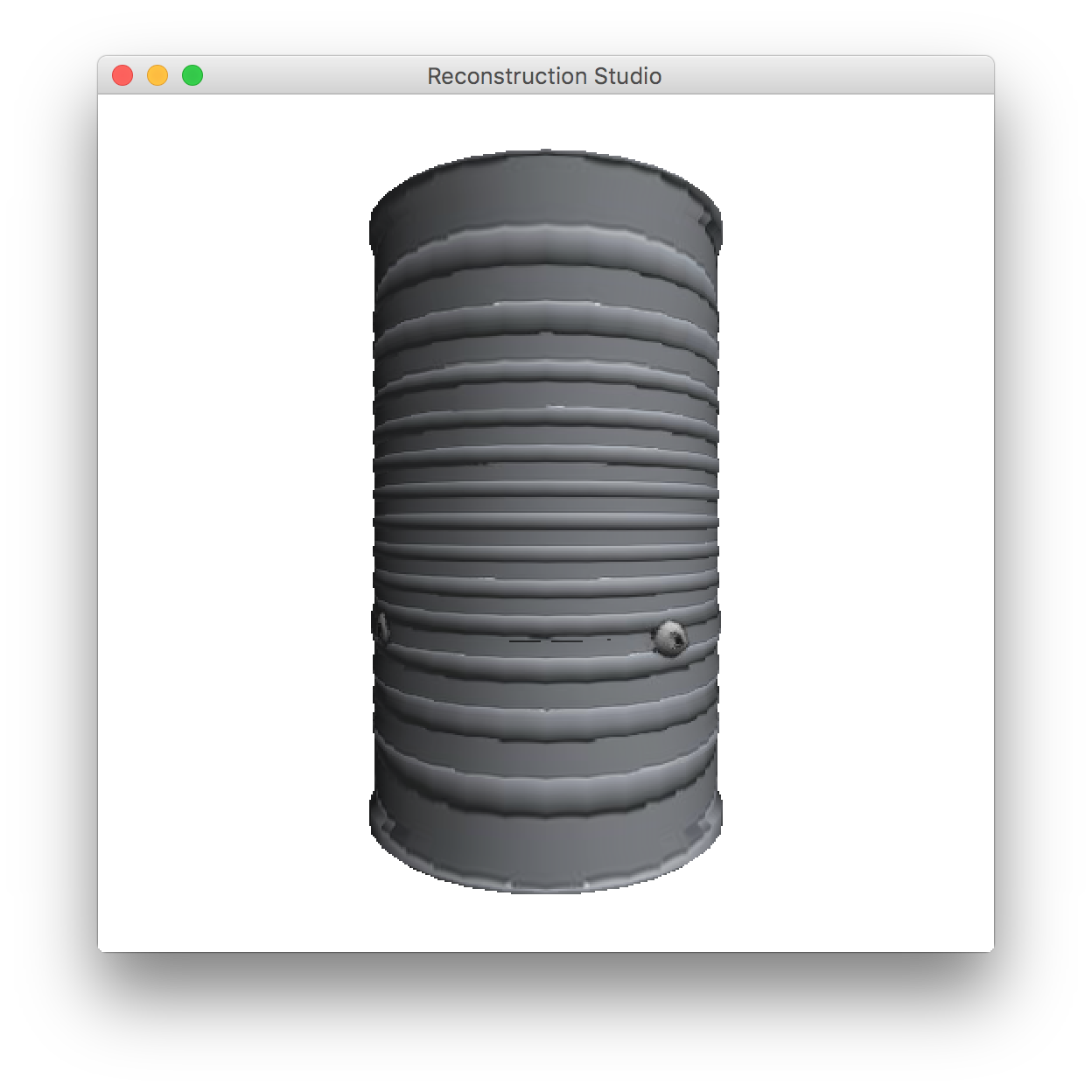}} \\
   \fbox{\includegraphics[trim={3cm 2.9cm 2.2cm 2cm},clip, width=0.19\linewidth]{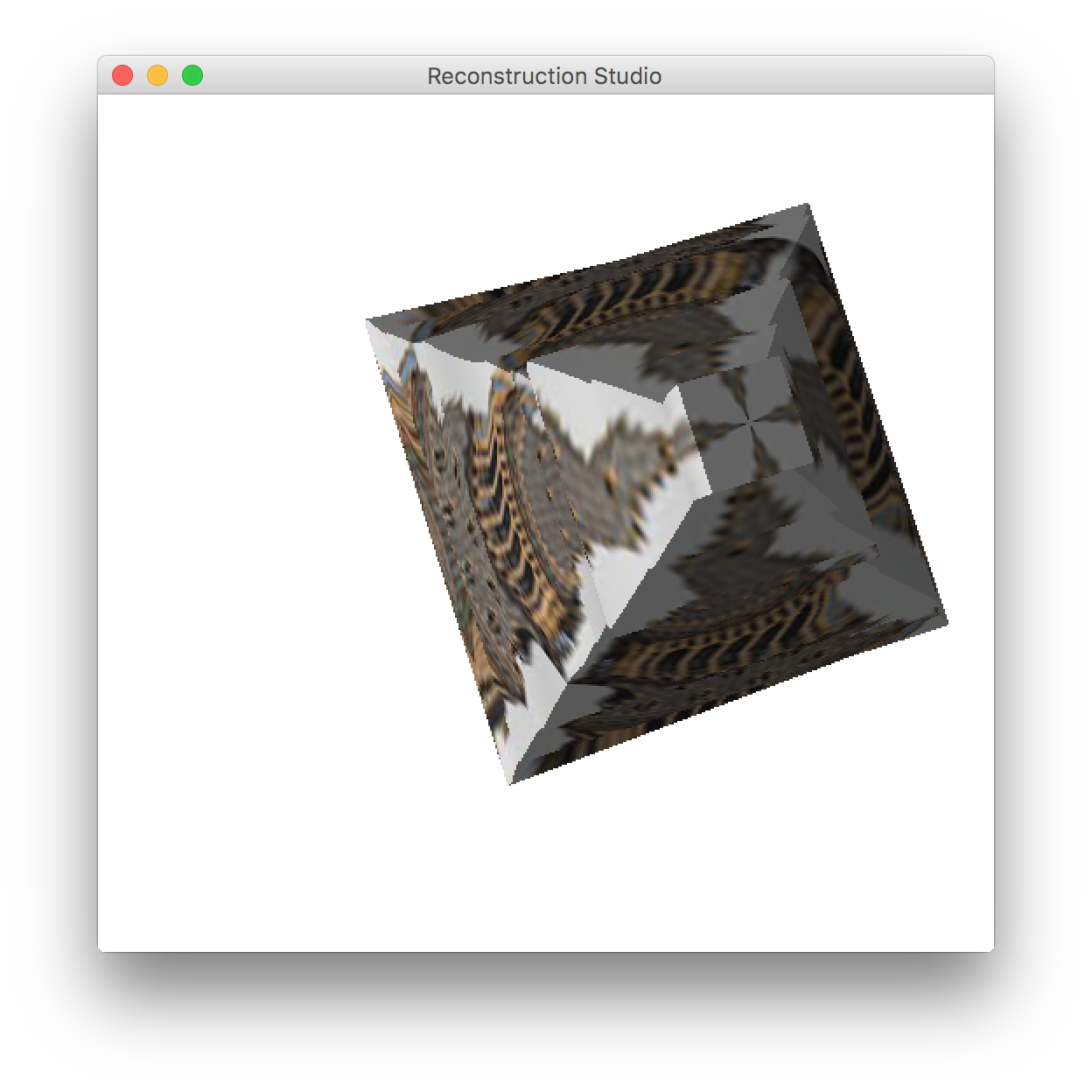}}
   \fbox{\includegraphics[trim={3cm 2.9cm 2.2cm 2cm},clip, width=0.19\linewidth]{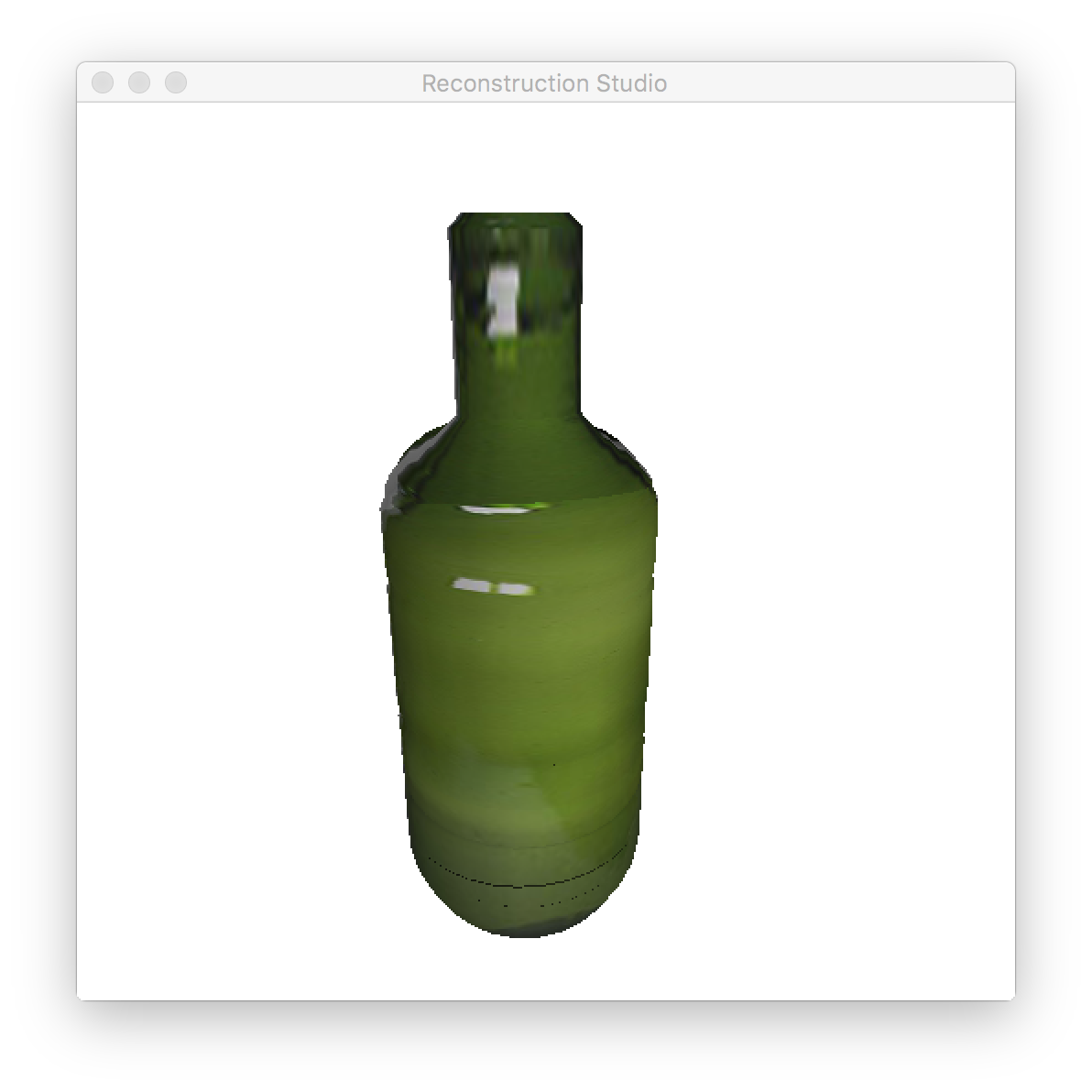}}
   \fbox{\includegraphics[trim={3cm 2.9cm 2.2cm 2cm},clip, width=0.19\linewidth]{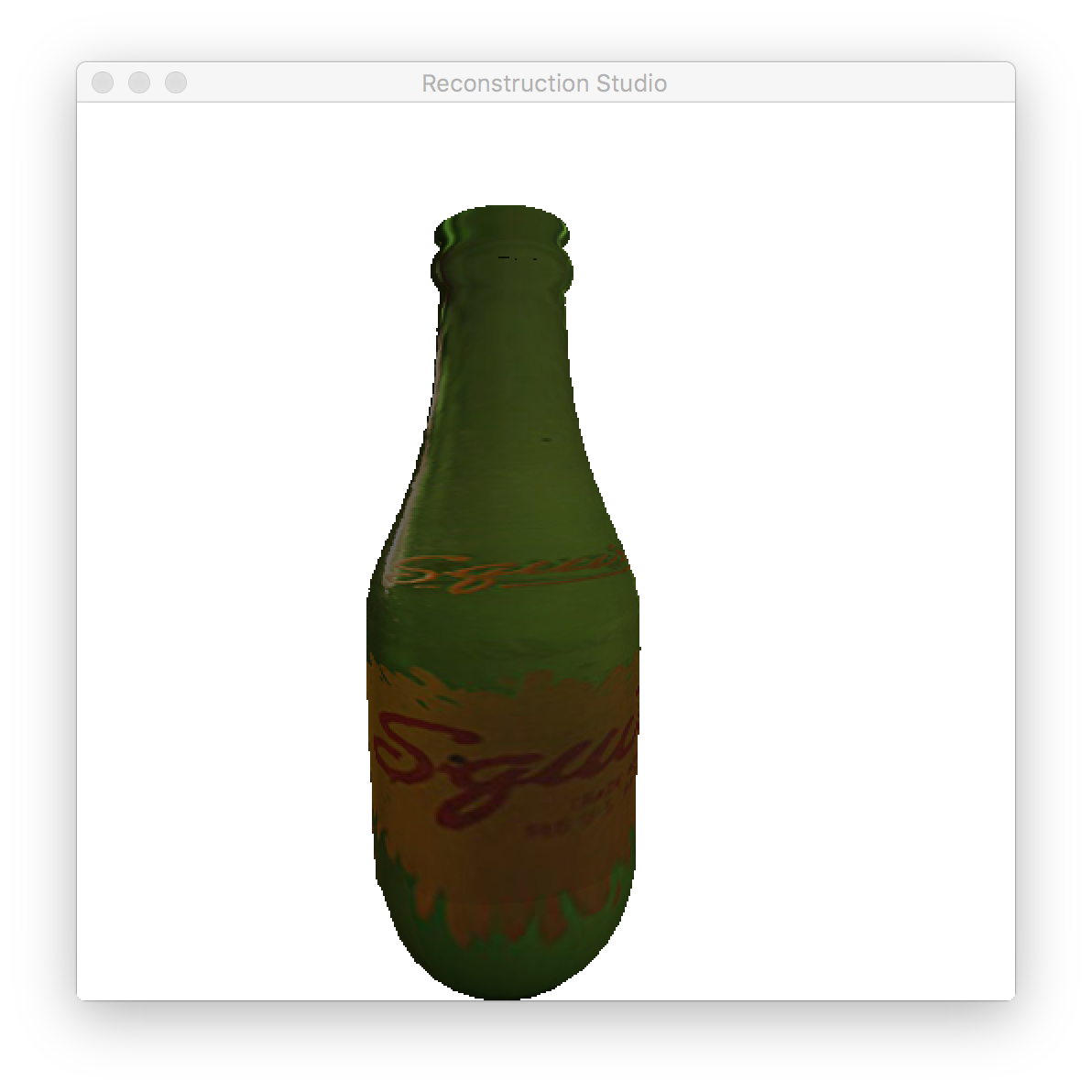}}
   \fbox{\includegraphics[trim={3cm 2.9cm 2.2cm 2cm},clip, width=0.19\linewidth]{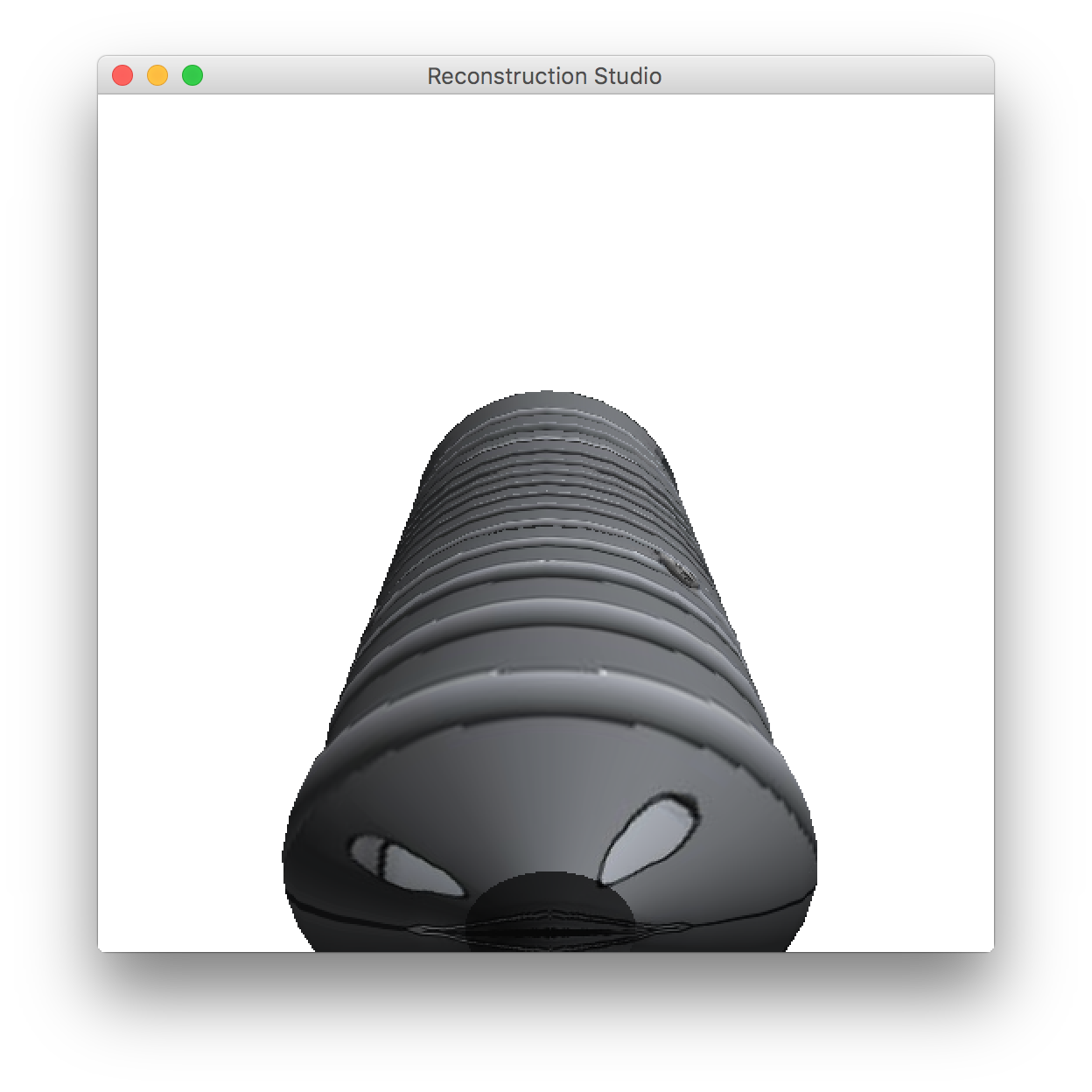}}
   \\ . \\
   \fbox{\includegraphics[width=0.19\linewidth]{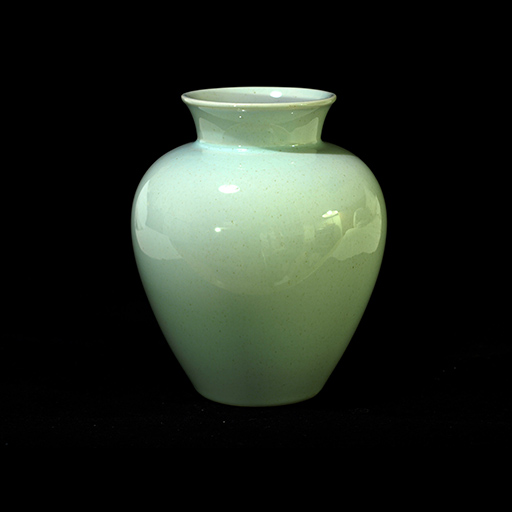}}
   \fbox{\includegraphics[width=0.19\linewidth]{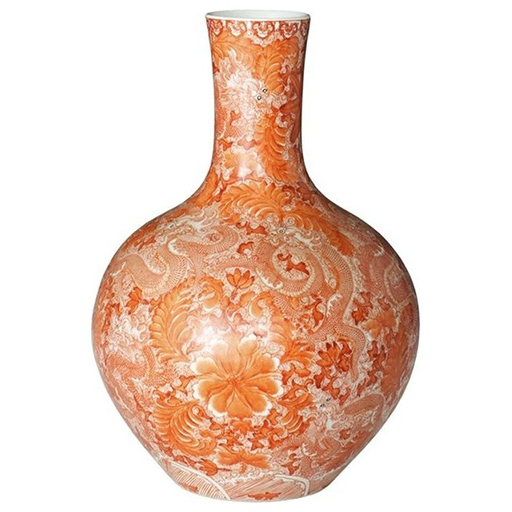}}
   \fbox{\includegraphics[width=0.19\linewidth]{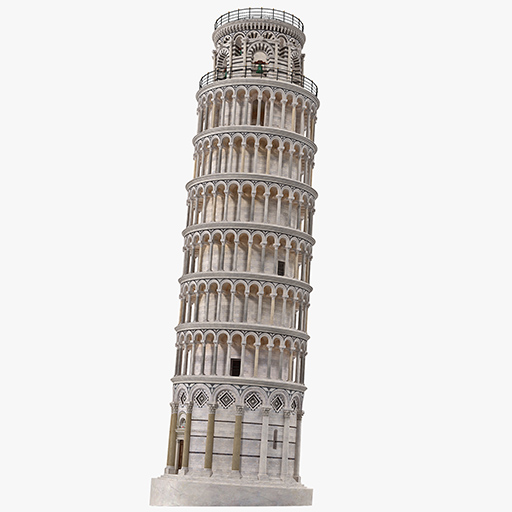}}
   \fbox{\includegraphics[width=0.19\linewidth]{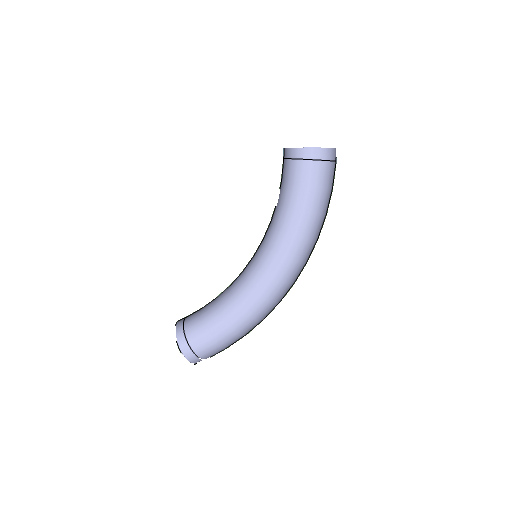}} \\
   \fbox{\includegraphics[trim={3cm 2.9cm 2.2cm 2cm}, clip, width=0.19\linewidth]{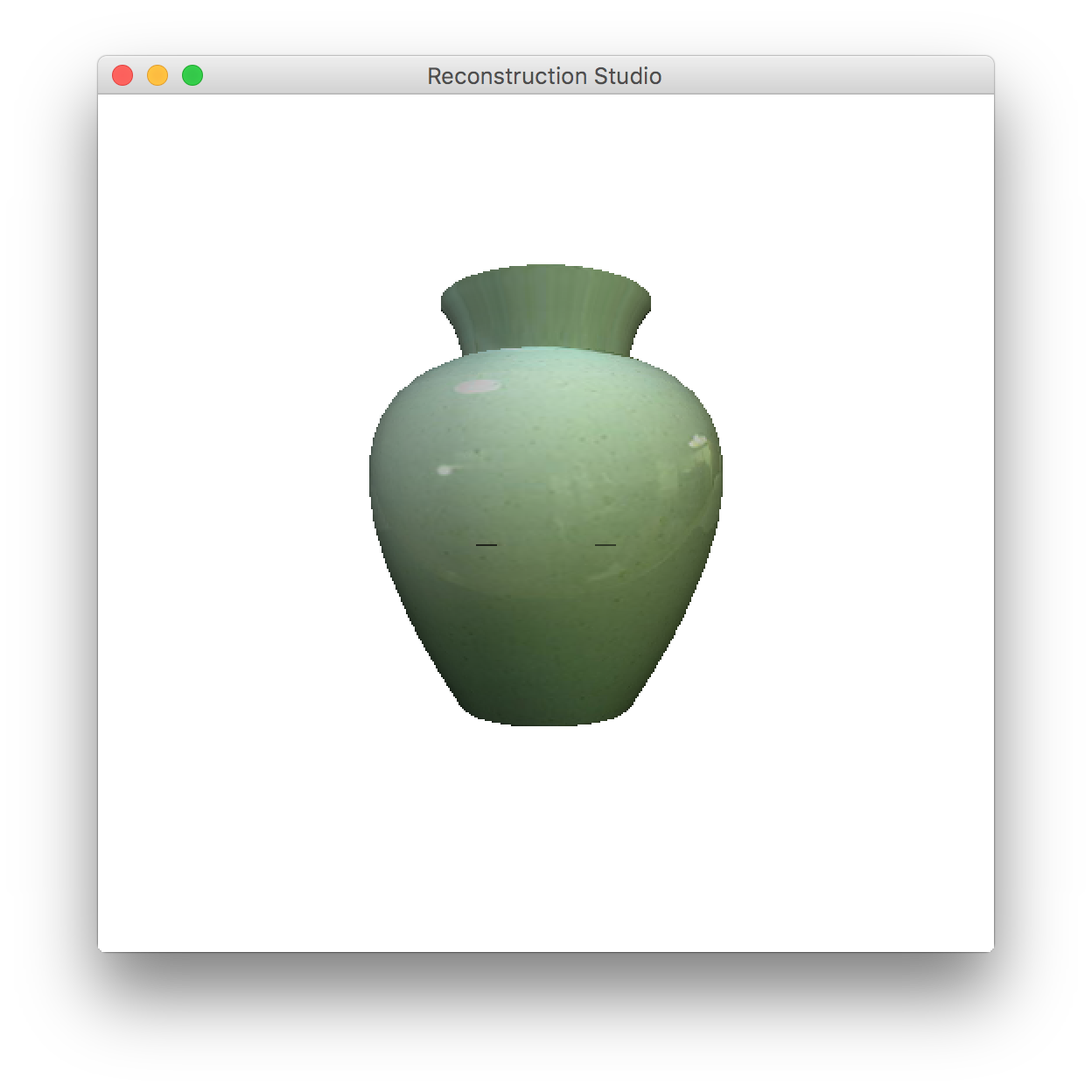}}
   \fbox{\includegraphics[trim={3cm 2.9cm 2.2cm 2cm}, clip, width=0.19\linewidth]{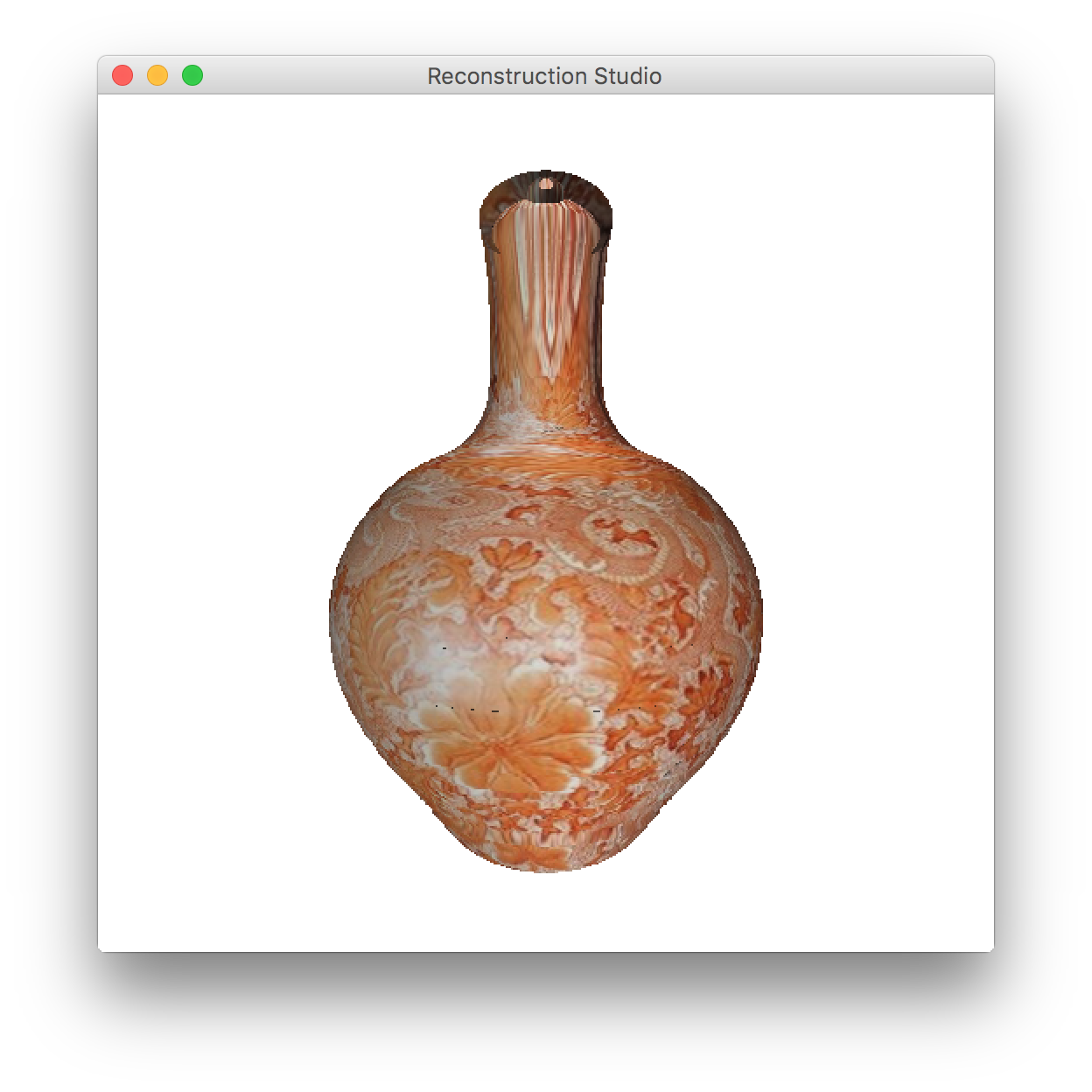}}
   \fbox{\includegraphics[trim={3cm 2.9cm 2.2cm 2cm}, clip, width=0.19\linewidth]{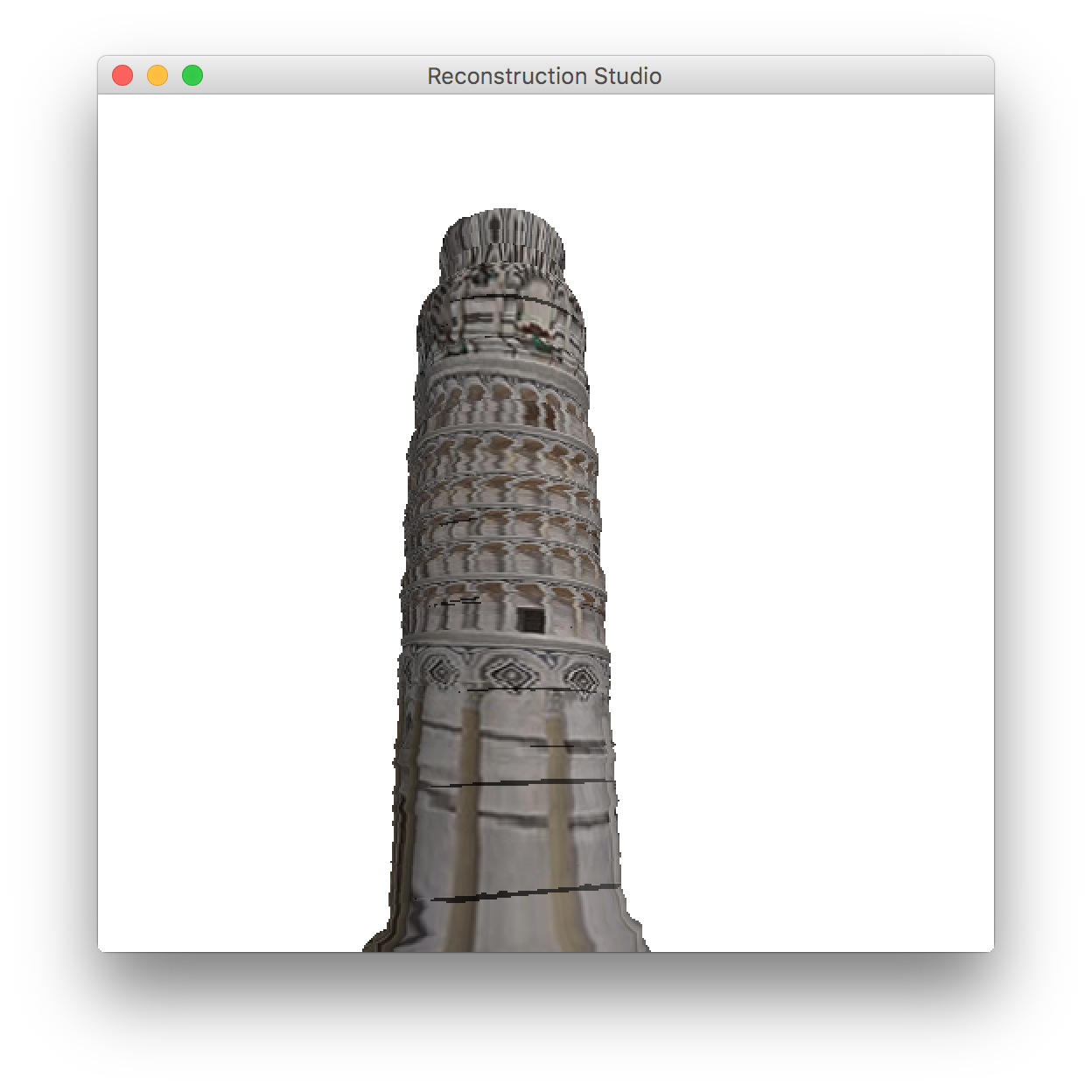}}
   \fbox{\includegraphics[trim={3cm 2.9cm 2.2cm 2cm}, clip, width=0.19\linewidth]{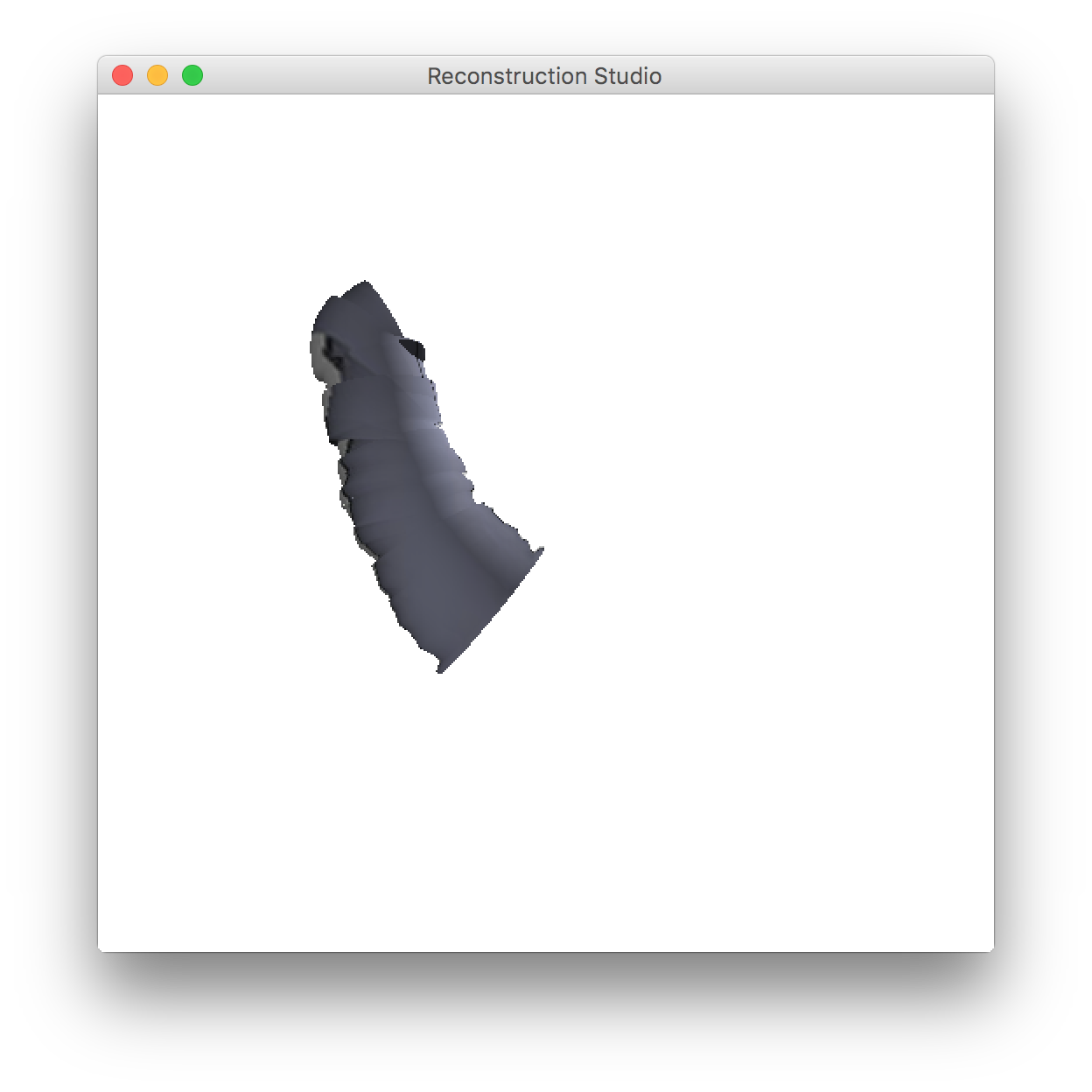}} \\
   \fbox{\includegraphics[trim={3cm 2.9cm 2.2cm 2cm}, clip, width=0.19\linewidth]{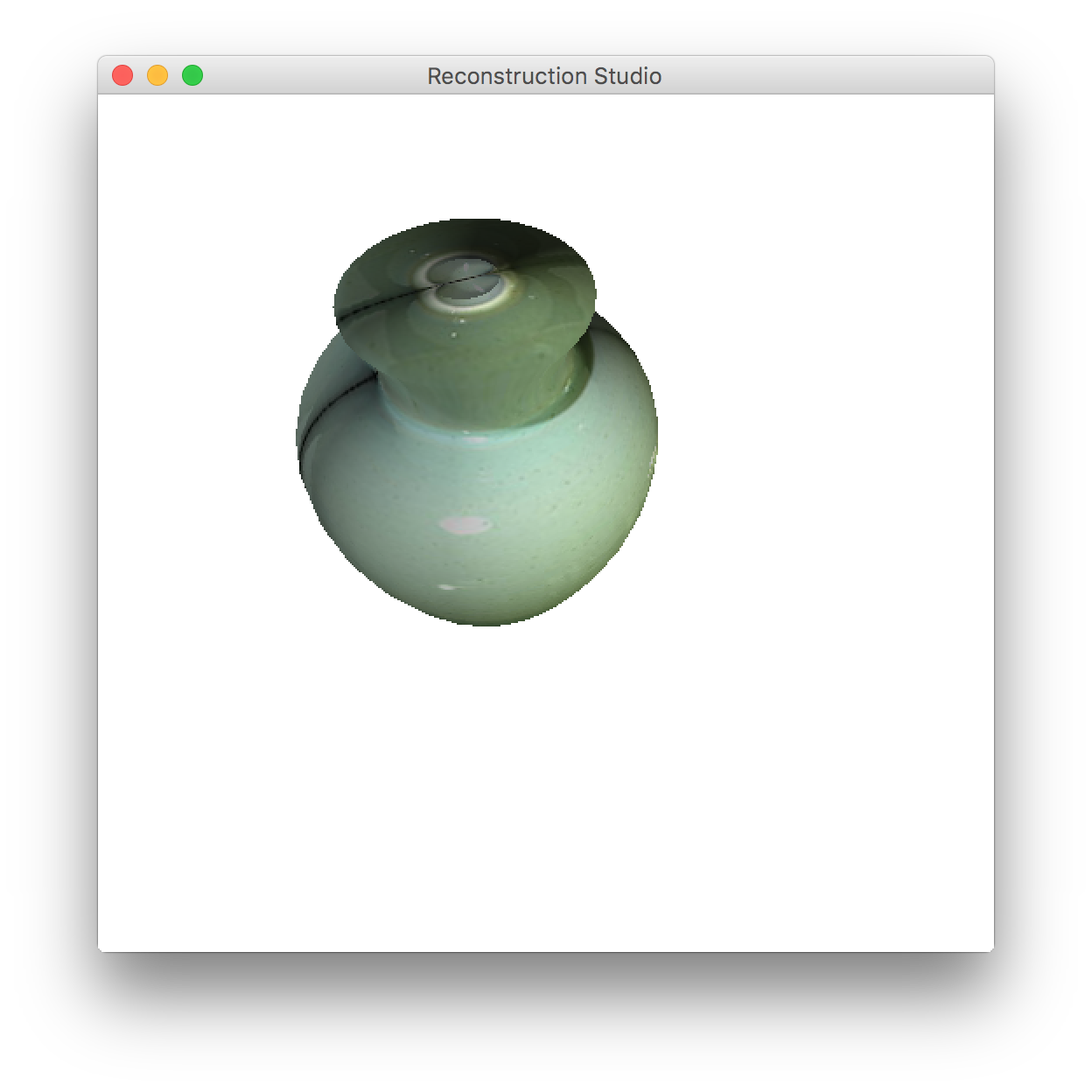}}
   \fbox{\includegraphics[trim={3cm 2.9cm 2.2cm 2cm}, clip, width=0.19\linewidth]{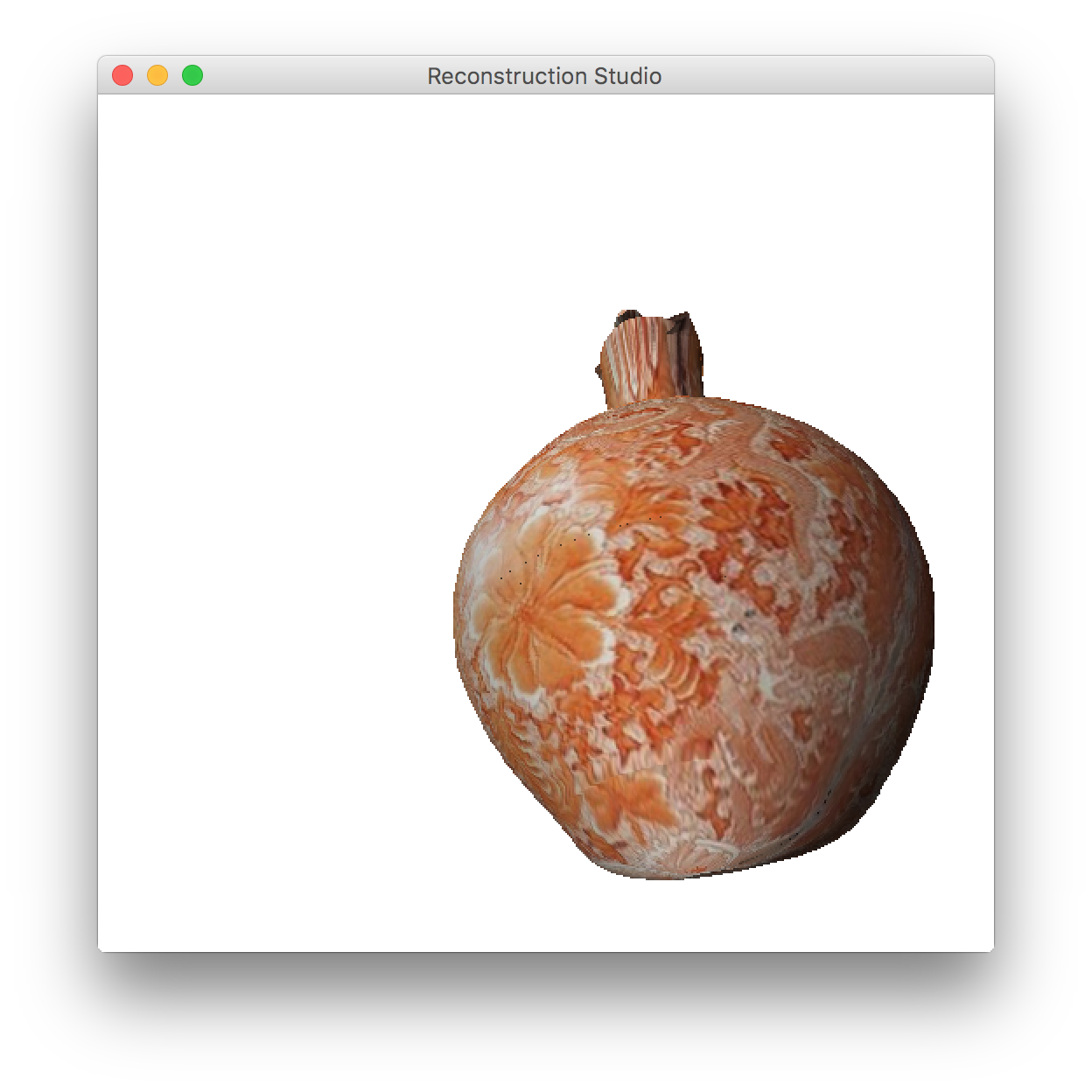}}
   \fbox{\includegraphics[trim={3cm 2.9cm 2.2cm 2cm}, clip, width=0.19\linewidth]{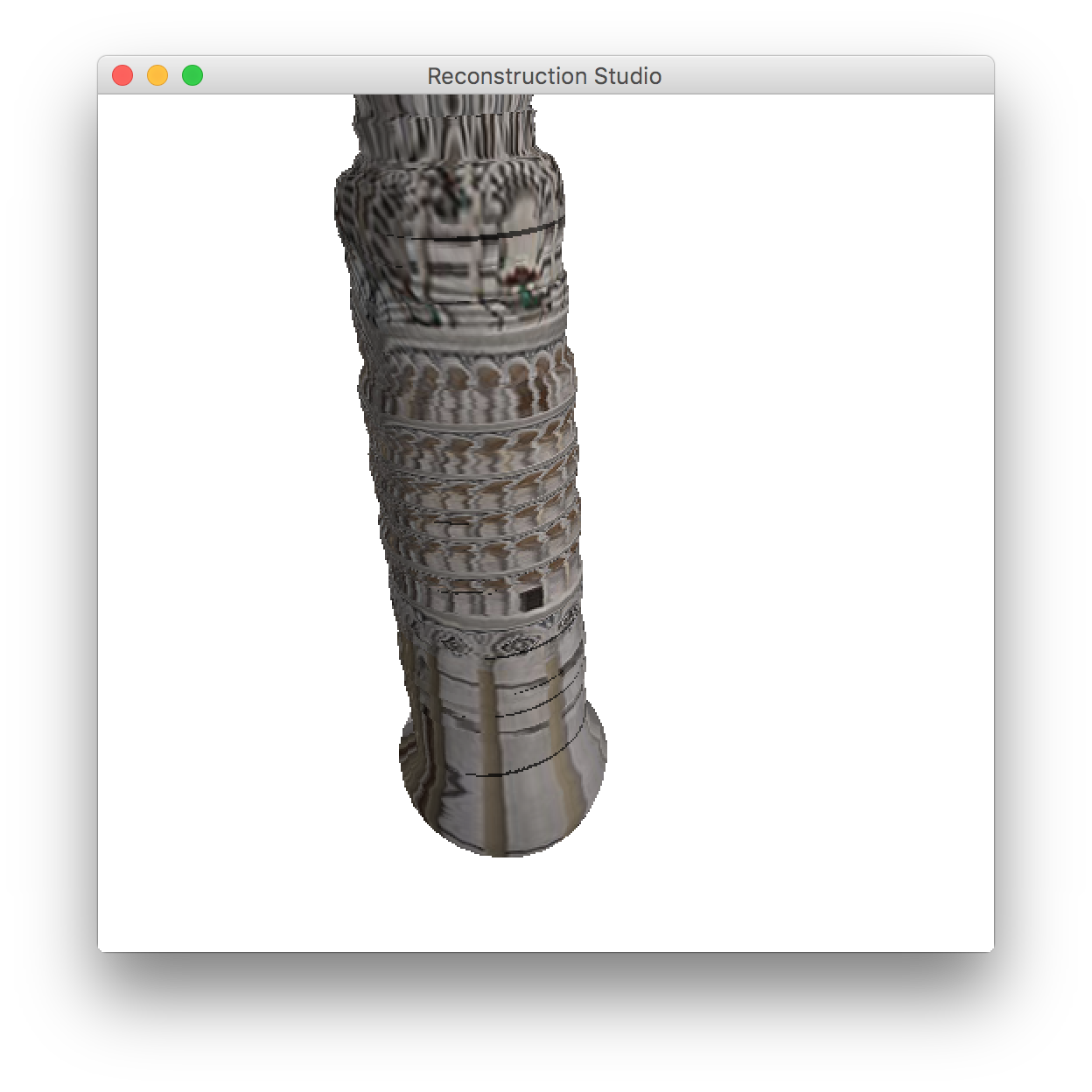}}
   \fbox{\includegraphics[trim={3cm 2.9cm 2.2cm 2cm}, clip, width=0.19\linewidth]{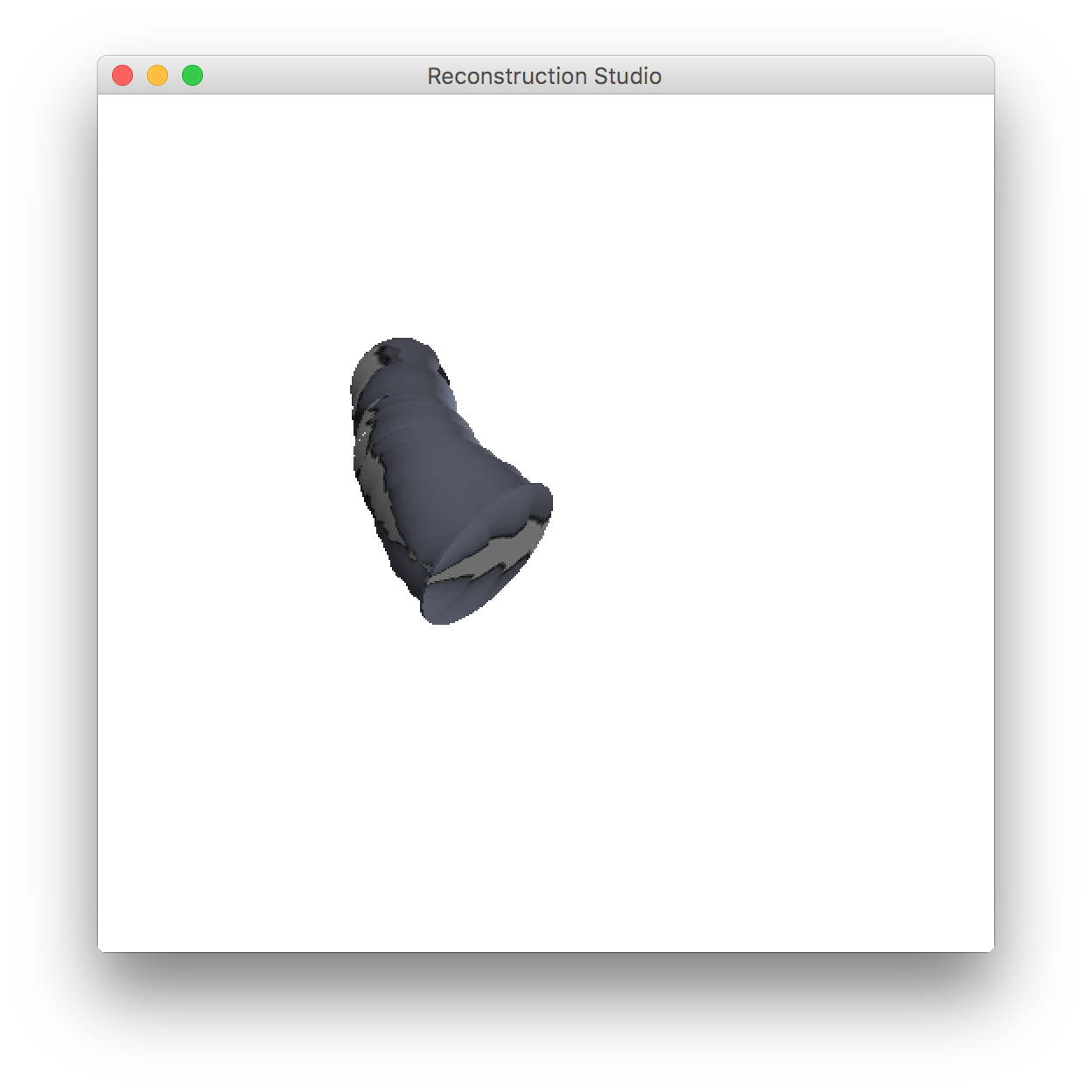}}

\end{center}
   \caption{Some sample input output sets. First row is the input while the next ones are outputs from two camera angles}
\label{fig:finaloutput}
\end{figure*}
\section{Results}
The process performs well as long as the constraints discussed during the process. Figure \ref{fig:finaloutput} gives some sample outputs. There are assumptions that directly affect the output. The process fails if:
\begin{itemize}
\item \textit{The object fails in segmentation:} We rely on segmentation to give a very accurate mask that can be used for regeneration. If grabcut fails to find a mask, we don't have any information about where the object is.
\item \textit{The object position is not a front view/with the extrusion surface on the top:} the direction of extrusion as well as the location of the surface of extrusion is assumed to be top-bottom.
\item \textit{Angle of the surface is not zero:} The surface is assumed to be perpendicular to the plane of the image. There is some robustness but the process can fail spectacularly if not.
\item \textit{The object is not smooth:} The Object is assumed to have a smooth surface with uniform transitions. Support for non-uniform transitions is an extension that is beyond the scope of the project.
\item \textit{Concavity} The concavity in the extrusion surface is supported but concavities during extrusion will be missed with most sweep based methods. A Workaround that tool can deploy to support concavities is subtraction of shapes from one another where the shape representing a hole can be generated manually and subtracted from the main shape.
\item \textit{The extrusion plane is not parallel to the image plane:} This assumption roots itself in the perspective ambiguity we have from a single image. This can be fixed with generating a set of models based on certain assumptions and picking the best fit. The output with such an image is shown in Figure \ref{fig:tilt}
\end{itemize}
\section{Conclusion and Future Work}
The scope if the current project and the set of inputs it supports is fairly restricted. But the overall field of sweep based modeling has a variety of use cases. Most of the human industrial output can be modeled by combining extruded shapes. 3D printing works on a principle of attaching layers of objects one over another, which is very similar to the concept of reconstruction used in this project. The outputs of an extrusion type modeling are suitable for 3D printing. \par
After removing some of the limitations via a learning system or using a better fitting technology, we can obtain a method that can solve reconstruction for a huge set of use cases. A fully automatic reconstruction engine has a lot of use cases in a variety of fields and reconstruction of extrusion based objects can be used directly as a piece of such a system.
{\small
\bibliographystyle{ieee}
\bibliography{egbib}
}

\end{document}